\documentclass[letterpaper,journal]{IEEEtran}
\usepackage{amsmath,amsfonts}
\usepackage{algorithmic}
\usepackage{algorithm}
\usepackage{array}
\usepackage[caption=false]{subfig}
\usepackage{textcomp}
\usepackage{stfloats}
\usepackage{url}
\usepackage{verbatim}
\usepackage{graphicx}
\usepackage{cite}
\usepackage{booktabs}
\usepackage[hidelinks]{hyperref}

\DeclareMathOperator*{\argmin}{arg\,min}
\begin{document}

\title{Compositional Entertainment Video Reconstruction}

\author{Sangmin~Kim,
        Seunguk~Do,
        Daeun~Lee,
        and~Jaesik~Park%
\thanks{
    Sangmin~Kim, Seunguk~Do, Daeun~Lee, and~Jaesik~Park are with Seoul National University, Republic of Korea.}
\thanks{
    Email:\texttt{\{sm.kim, seunguk.do, del0716, jaesik.park\}}\allowbreak\texttt{@snu.ac.kr}}
\thanks{
    Corresponding author: Jaesik Park.}
}

\markboth{preprint}
{Kim \MakeLowercase{\textit{et al.}}: Compositional Entertainment Video Reconstruction}


\maketitle
\begin{abstract}
Reconstructing dynamic radiance fields from video clips is challenging, especially when entertainment videos like TV shows are given. Many challenges make the reconstruction difficult due to actors occluding with each other and having diverse facial expressions, cluttered stages, and small baseline views or sudden shot changes. 
Reconstruction becomes even more challenging when dealing with general monocular web videos, which present an even greater degree of unpredictability and complexity compared to controlled environments.
To address these issues, we present ShowMak3r++, a unified reconstruction pipeline that targets both controlled settings like TV shows and uncontrolled scenarios like web videos.
Our pipeline allows the editing of scenes like how video clips are made in a production control room after the reconstruction is done. 
In our pipeline, we propose a spatio-temporal positioning module that locates actors on the stage by using depth prior while maintaining 2D image alignment and natural 3D motions.  
ShotMatcher module then tracks the actors under shot changes. Finally, a face-fitting network dynamically recovers the actors’ expressions. 
Experiments on Sitcoms3D and CMU Panoptic datasets show that our pipeline can reassemble TV show scenes with new cameras at different timestamps. 
We also demonstrate that our method can successfully reconstruct challenging web videos, including dynamic action clips, dance videos, and movie clips.
Furthermore, we demonstrate that our pipeline enables interesting applications such as synthetic shot-making, actor relocation, insertion, deletion, and pose manipulation. Project page: https://nstar1125.github.io/showmak3r/.
\end{abstract}


\section{Introduction}\label{sec:introduction}
\IEEEPARstart{C}{onsiderable} advances in radiance field reconstruction approaches~\cite{mildenhall2021nerf, kerbl20233d} transform how we reconstruct and visualize the scenes. Recent methods aim to bring video clips into 4D space to enable novel viewpoint rendering or scene editing. However, recovering the radiance field from dynamic scenes remains a challenging problem. Approaches in this category have mainly focused on scenarios with multi-view synchronized cameras or fully observed scenes~\cite{park2021hypernerf, jung2023deformable, wu20244d}.

The reconstruction gets even harder for entertainment videos, such as TV shows captured by shot-changing (video transition by another camera) monocular cameras. Compared to existing benchmark datasets~\cite{jiang2022neuman}, entertainment videos like TV shows present additional challenges. First, it contains scenes that are inherently hard to reconstruct, such as multiple actors interacting and occluding each other on the cluttered stages or actors showing detailed facial changes to express their emotions. In addition, the videos are filmed with multiple cameras and then edited to appear as a continuous timeline, resulting in sudden shot changes. Furthermore, cameras are mainly positioned in front of the scene, creating partial observations and thus limiting information about the actors' backsides. Therefore, even the state-of-the-art methods~\cite{kocabas2024hugs, som2024} fail to recover consistent dynamic radiance fields due to incorrect human-scene alignment and inconsistent deformation of human movements.
Recent data-driven approaches~\cite{zhang2024monst3r, lu2025align3r, wang2025cut3r, li2025megasam} show notable performance on reconstructing from monocular web video inputs. However, they focus on estimating point clouds or human meshes, which are not photo-realistic, thus making them unsuitable for post-production.

\begin{figure}[t]
    \centering
    \includegraphics[width=0.5\textwidth]{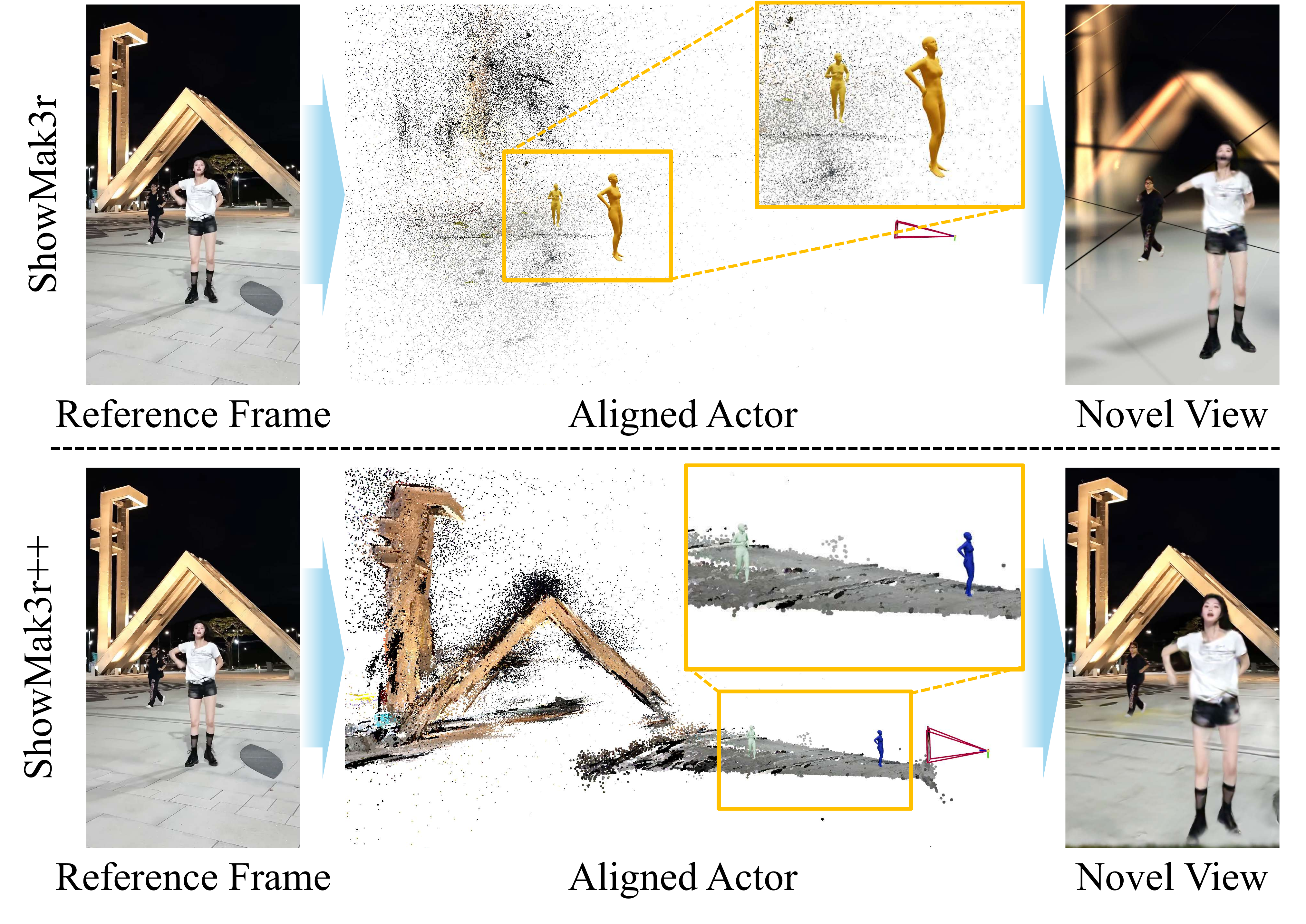}
    \vspace{-8mm} 
    \caption{
    We present ShowMak3r++, a compositional video reconstruction pipeline that positions the actor to a more accurate location and targets more diverse scenarios than our previous work, ShowMak3r.
    }
    \vspace{-2mm}
    \label{fig:intro}
\end{figure}

In this work, we present a comprehensive total reconstruction pipeline that reconstructs the dynamic radiance field from entertainment videos, enabling editing like how the video clip is made in a production control room. 
We first build a 3D stage and recover parametric models of dynamic actors. Since the estimated humans from the video clip and the reconstructed stage have different coordinate systems, 
we propose \textit{spatio-temporal positioning} module, which aligns the actors to their correct locations on the stage by leveraging depth estimation while maintaining 2D image alignment and natural 3D motions.
\textit{Spatio-temporal positioning} module also aims to solve unseen human poses via interpolation. 
After positioning the actors, we use the \textit{ShotMatcher} module to perform human association at the shot boundaries to track actors across different shots.
As the facial change of the actors is a key element in entertainment videos, we implement an implicit face-fitting network to change human expressions across frames to address this dynamically. 

This paper is based on our previous work, ShowMak3r~\cite{kim2025showmak3r}, which addresses several key limitations. 
First, ShowMak3r is designed for controlled environments like TV Shows, where aggregating additional background images from different episodes is possible. 
However, this assumption limits its use on general monocular web videos. 
Furthermore, ShowMak3r uses a module called `3DLocator' which relies solely on aligned depth maps for positioning actors. This results in problems such as parts of the actors intersecting with the 3D stage or actors exhibiting unnatural jittering movements in 3D space.

To address these issues, we propose a new pipeline, ShowMak3r++, which is an extension of our previous work, ShowMak3r. Unlike our previous work, which only targets controlled environments like TV Shows, ShowMak3r++ can successfully reconstruct dynamic scenes from uncontrolled web video environments without relying on additional background images. 
We also propose our new \textit{spatio-temporal positioning} module that aligns actors to more precise locations by considering 3D trajectories and stage penetrations. 

Experiments with the Sitcom3D and Panoptic dataset~\cite{pavlakos2022one, joo2015panoptic} show that our pipeline can successfully reconstruct from entertainment videos like TV shows. 
We also compare qualitative results on various web video scenarios, such as dynamic action clips, dance videos, or movie clips, with other web video reconstruction works, proving our method's validity.
Furthermore, we demonstrate various applications, including actor relocation, insertion, deletion, and pose manipulations. 

Our contributions can be summarized as below:
\begin{itemize}
\item  We introduce a comprehensive pipeline that reconstructs the dynamic radiance field from entertainment videos, enabling editing like a production control room.
\item Our method can be applied to both controlled settings, like TV shows, and uncontrolled web video scenarios, like dynamic action clips, dance videos, or movie clips.
\item We propose \textit{spatio-temporal positioning} module that accurately aligns actors on the stage while maintaining 2D image alignment and natural 3D motions.
It even aims to solve unseen poses due to occlusions.
\item We present \textit{ShotMatcher} that enables continuous tracking of actors under shot change. Our approach associates actors even when they are not visible in certain shots.
\item We implement an implicit \textit{face-fitting network} to recover and express dynamic facial expressions.
\item Extensive experiments show the validity of our approach and demonstrate possible applications such as actor relocation, insertion, deletion, and pose manipulation.
\end{itemize}
\section{Related Work}
\textbf{4D Scene Reconstruction.}
Attempts have been made to extend radiance fields~\cite{mildenhall2021nerf, kerbl20233d} into spatio-temporal models~\cite{li2022neural, park2021hypernerf, li2023dynibar, park2021nerfies, pumarola2021d, cao2023hexplane, jung2023deformable, wu20244d} for reconstructing 4D scenes from synchronized multi-camera videos. These techniques have since evolved to handle single-view video inputs~\cite{Li_2021_CVPR, li2023dynibar, lei2024mosca, stearns2024dynamic, wang2024gflow, som2024, Fridovich-Keil_2023_CVPR, song2023nerfplayer, lin2024gaussian, lu2025bardgs}, increasing their applicability across various scenarios. More recently, feed-forward methods that directly generate dynamic point clouds have emerged~\cite{zhang2024monst3r, lu2025align3r, wang2025cut3r, wang2025mem4d}, eliminating the need for per-scene training. However, despite these advancements, current 4D reconstruction methods still struggle with TV show content, which often features limited camera angles, rapid human motion, and abrupt scene transitions.

\vspace{2mm}\noindent\textbf{3D Avatar Reconstruction from Videos.} Human radiance fields~\cite{weng2022humannerf, zhao2022humannerf, guo2023vid2avatar, yu2023monohuman, jiang2022neuman, geng2023learning, hu2024gaussianavatar, hu2024gauhuman, jiang2024hifi4g, kocabas2024hugs, li2024animatable, moon2024expressive, moreau2024human, pang2024ash, qian2024gaussianavatars, qian20243dgsavatar, shao2025degas, shao2024splattingavatar, isik2023humanrf, su2021nerf, peng2021animatable, buehler2025dla, dong2025moga, guo2025vid2avatarpro}, which incorporates radiance fields~\cite{mildenhall2021nerf, kerbl20233d} with parametric human models~\cite{loper2023smpl, pavlakos2019expressive}, have demonstrated that photo-realistic 3D human avatars can be reconstructed from monocular video. These representations later evolved into generalizable humans radiance fields~\cite{hu2023sherf, li2024ghunerf, dey2024ghnerf, wang2024eg, mu2023actorsnerf, yu25gaia}, which eliminated the need for extensive per-avatar training time. More recently, approaches to adopt diffusion models~\cite{rombach2022high, podell2023sdxl} as additional priors~\cite{wang2023prolificdreamer, jain2022zero, poole2022dreamfusion, graikos2022diffusion} for recovering unseen regions have been studied~\cite{lee2024guess, dutta2025chrome}.
When reconstructing 3D humans from TV shows, facial features are particularly crucial. Researchers have proposed utilizing pretrained expression encoders~\cite{shao2025degas} or directly leveraging SMPL-X expression parameters\cite{moon2024expressive} to enhance facial features. However, these approaches typically demand multi-view face images~\cite{shao2025degas, aneja2025scaffoldavatar} or face challenges in capturing nuanced expressions from TV show videos~\cite{moon2024expressive}. To address these limitations, we propose a simple, yet highly effective approach of refining facial expressions through an implicit deformation network.

\vspace{2mm}\noindent\textbf{Composite Human-Scene Reconstruction.} Previous approaches for reconstructing scenes with humans~\cite{li2022neural, kocabas2024hugs, guo2023vid2avatar, guo2025vid2avatarpro} treat backgrounds as static and use separate representations for humans and backgrounds. However, these methods require videos to capture the entire human body, including foot contact with the floor, making them unsuitable for TV show settings.
Sitcoms3D~\cite{pavlakos2022one} addresses it by reconstructing sitcom videos through NeRF-W~\cite{martin2021nerf} for consistent backgrounds and optimizing SMPL parameters using adjacent shots. However, it lacks human texture and requires identical humans to appear in neighboring shots.
OmniRe~\cite{chen2024omnire} reconstructs outdoor scenes with dynamic objects, including pedestrians and vehicles. However, it relies on LiDAR sensors for geometric data and isn't designed for scenarios with shot changes.
In contrast, our method effectively positions actors within the stage without requiring multiple shots, foot contact points, or additional sensors.
\begin{figure*}[ht]
    \centering
    \includegraphics[width=0.91\textwidth]{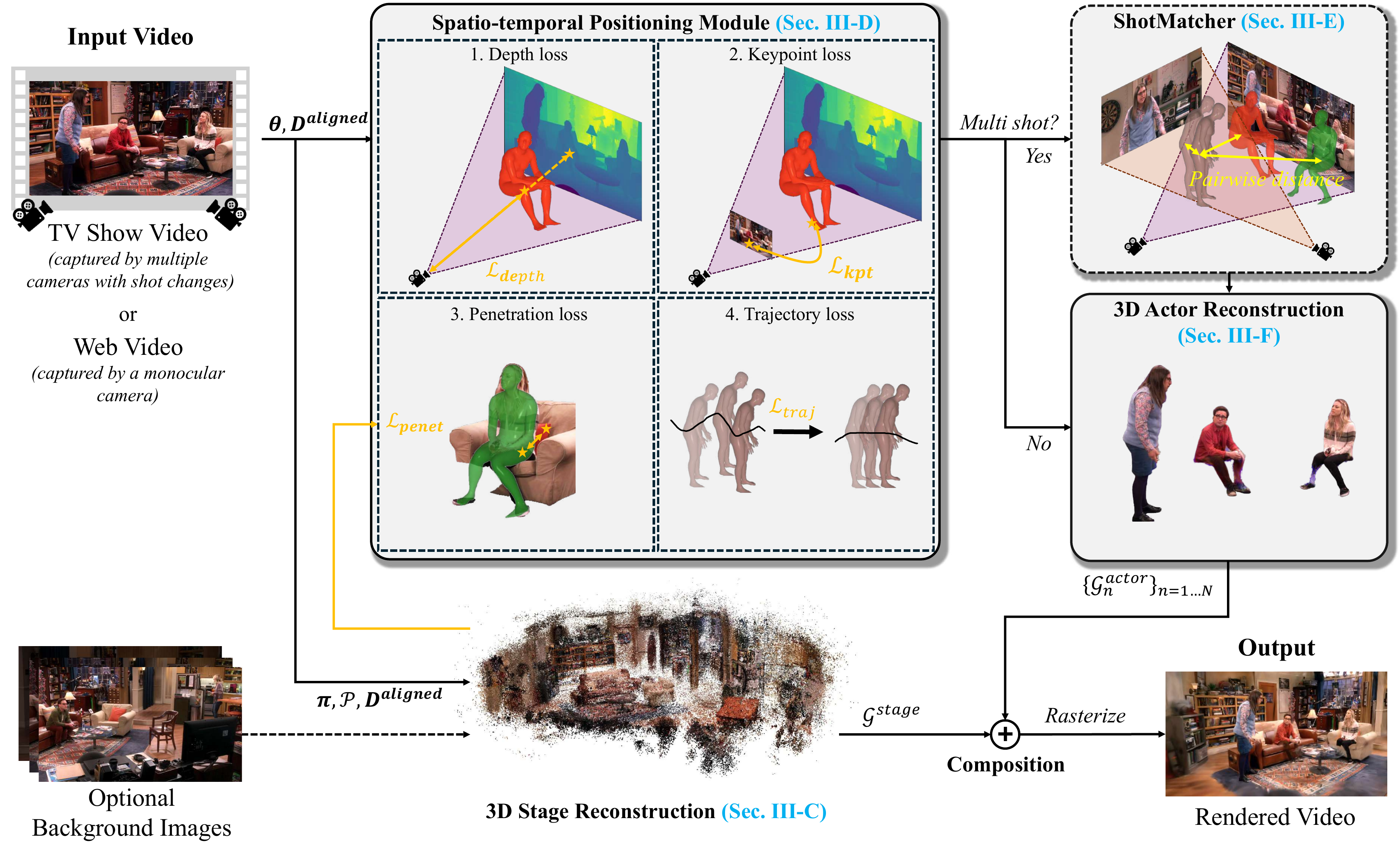}
    \vspace{-3mm}
    \caption{Overview of our ShowMak3r++ pipeline. Given an entertainment video clip, we perform dense reconstruction of the stage (Sec.~\ref{sec:3D_stage}), locate SMPL models to the stage while maintaining 2D image alignment and natural 3D motions (Sec.~\ref{sec:positioning}), and if more than a single shot is given, we associate SMPL models across the shots by calculating pairwise distance between actors at shot-boundary (Sec.~\ref{sec:tracking}). Then we recover the detailed appearance of the actors (Sec.~\ref{sec:3D_actor}). Finally, 3D Gaussians of the stage and the actors are rendered to produce novel frames.}
    \vspace{-2mm}
    \label{fig:pipeline}
\end{figure*}

\section{Method}
\subsection{Overview}
In entertainment videos like TV shows, several \emph{actors} perform on the \emph{stage}. The TV shows are structured into three hierarchical level semantics: \textit{scene}, \textit{shot}, and \textit{frame}. A scene represents a sequence of related shots that follow a continuous narrative flow~\cite{katz1979ephraim}, a shot captures continuous frames within a single camera~\cite{sklar1990film}, and a frame refers to an individual image within the sequence. 
Our work aims to reconstruct dynamic scenes from such frame hierarchy in TV shows. 
Furthermore, we extend our pipeline to be applicable for web video scenarios with uncontrolled environments.

As shown in Fig.~\ref{fig:pipeline}, Sec~\ref{sec:3D_stage} presents how we reconstruct the consistent stage. Sec~\ref{sec:positioning} introduces how our \textit{spatio-temporal positioning} module positions multiple actors into correct locations on the stage 
while maintaining 2D image alignment and 3D natural motions. Sec~\ref{sec:tracking} explains how our \textit{ShotMatcher} tracks actors across shot changes.
Sec~\ref{sec:3D_actor} shows how we reconstruct dynamic actors and their expressions with our \textit{face-fitting network}.

\vspace{2mm} \noindent\textbf{Scene representation.}
Our approach represents the stage and the actors with 3DGS~\cite{kerbl20233d}, an explicit approach that reconstructs a radiance field with 3D Gaussians. Each Gaussian consists of attributes, including center $\boldsymbol{\mu} \in \mathbb{R}^3$, rotation $\textbf{q} \in \mathbb{R}^4$, scale $\textbf{s} \in \mathbb{R}^3$, color $\textbf{c} \in \mathbb{R}^3$, and opacity $o \in \mathbb{R}$. The \textit{k}-th Gaussian is defined as follows:
\begin{equation}
g_k(\textbf{p})=o_k\text{exp}^{-\frac{1}{2}(\textbf{p}-\boldsymbol{\mu}_k)^T\boldsymbol{\Sigma}_k^{-1}(\textbf{p}-\boldsymbol{\mu}_k)},
\end{equation}
where position is $\textbf{p} \in \mathbb{R}^3$, and the covariance matrix is $\boldsymbol{\Sigma}_k = \textbf{R}_k\textbf{S}_k\textbf{S}_k^T\textbf{R}_k^T$. $\textbf{R}_k \in SO(3)$ and $\textbf{S}_k \in \mathbb{R}_+^3$ are obtained from quaternion of $\textbf{q}$ and scale $\textbf{s}$. Unlike implicit methods~\cite{jiang2022neuman}, 3D Gaussians $\mathcal{G}=\{g_k\}_{k=1...K}$ have an explicit nature, which makes it effective for reconstructing the stage $\mathcal{G}^{\text{stage}}$ and multiple actors $\mathcal{G}_n^{\text{actor}}$ by simply compositing multiple Gaussian sets as $\mathcal{G}^{\text{composite}} = \mathcal{G}^{\text{stage}} \cup \{\mathcal{G}_n^{\text{actor}}\}_{n=1...N}$.


\subsection{Preprocessing}\label{sec:preprocessing}
\textbf{Camera parameters from TV Shows.}
In controlled environments like TV shows, we utilize additional background images aggregated from other episodes. These images play a critical role in estimating accurate camera poses and the 3D structure. 
To recover the camera pose $\pi_f$ of each frame $f$ in entertainment videos like TV shows, we use Structure-from-Motion systems. We observe that GLOMAP~\cite{pan2024global} robustly handles panning frames by globally estimating camera poses and the 3D structure of all input images at once. To reduce the effect of transient actors, we mask input images with the binary segmentation map of actors $M_f^{actor}$ using SAM~\cite{kirillov2023segment}, and obtain $M_f^{stage}$ by inverting $M_f^{actor}$.

\vspace{2mm}\noindent\textbf{Camera parameters from web videos.}
For uncontrolled environments like web videos, additional background images are often unavailable. This leads to traditional Structure-from-Motion systems~\cite{schonberger2016structure, pan2024global} producing inaccurate results. To address this issue, we utilize data-driven methods such as Pi3~\cite{wang2025pi3} to acquire the precise camera poses and 3D structure of web videos. To reduce the memory size, we sample \textit{K} random points per frame. Furthermore, since we only need a point cloud of the stage, we mask out actors by using a binary segmentation map $M_f^{actor}$.

\vspace{2mm}\noindent \textbf{Guiding depth maps.}
By following the convention, we reconstruct $\mathcal{G}^{\text{stage}}$ using 3DGS~\cite{kerbl20233d} given the camera poses $\pi$. However, we observed that vanilla 3DGS struggles with the reconstruction due to the partially observed backgrounds or narrow baselines, which are frequent in entertainment videos. Therefore, we utilize the monocular depth as guidance to compensate the limited observations. After we get dense depth predictions $\{D^{\text{mono}}_f\in \mathbb{R}^{H \times W} | f = 1, 2, ..., F\}$ from a data-driven approach~\cite{wang2025moge}, we adjust scale $a$ and offset $b$ of each depth map to match the scale of the predicted camera coordinate system.

Specifically, given the 3D point clouds $\mathcal{P}_f$ visible at $f$-th frame, we find $a^*$ and $b^*$ as follows:
\begin{equation}
a^*, b^* =\argmin_{a,b} \sum_{{\mathbf{p}}\in \mathcal{P}_f} \mathcal{L}(\textbf{p}; a, b),
\end{equation}
where the Huber loss $\mathcal{L}(p; a, b)$ with the depth of projected point $p_z$ for a view $\pi$ is defined as follows:
\begin{equation}
\mathcal{L}(\textbf{p}; a, b) \!=\!
\begin{cases} 
\frac{1}{2} (p_z \!-\! (aD^{\text{mono}}(\pi(\textbf{p}))\!+\!b))^2 & \hspace{-0.7cm} \text{if } |\gamma_z| \leq \delta_1, \\
\delta_1 (|p_z \!-\! (aD^{\text{mono}}(\pi(\textbf{p}))\!+\!b)| \!-\! \frac{\delta_1}{2}) & \hspace{-0.2cm} \text{otherwise}
\end{cases}
\label{eq:huber_loss}
\end{equation}
where we empirically set $\delta_1 = r_{\text{stage}}/100$, where $r_\text{stage}$ represents the scene radius. $\gamma_z$ denotes $p_z \!-\!(aD^{\text{mono}}(\pi(\textbf{p}))\!+\!b)$. We then obtain the depth map aligned with the camera coordinate system by calculating $D^{\text{aligned}} = a^* \times D^{\text{mono}}+ b^*$. We iterate the above process for the entire frame.

Note that the aligned depth map $D^{\text{aligned}}$ is a key component of our pipeline that boosts stage reconstruction (Sec.~\ref{sec:3D_stage}) and guides the positioning of the actors (Sec.~\ref{sec:positioning}). 


\subsection{3D Stage Reconstruction}\label{sec:3D_stage}
We reconstruct dense Gaussians of the stage $\mathcal{G}^{\text{stage}}$ using 3DGS~\cite{kerbl20233d} with the extended loss that leverages the aligned depth maps $D^\text{aligned}$ obtained from Sec.~\ref{sec:preprocessing}. We observe that depth guidance provides denser and more reliable reconstruction of the stage compared to using only photometric loss.

\vspace{2mm}\noindent\textbf{Background images.}
To recover a complete stage, we can leverage optional background images. For example, entertainment videos like sitcoms depict a similar environment over the season, so shots in various episodes show diverse views of the stages, which we can use for the stage reconstruction~\cite{pavlakos2022one}.
While utilizing additional background images yields a more complete stage, our method remains robust even when they are unavailable, as in web video scenarios.

\vspace{2mm}\noindent\textbf{Depth-guided dense reconstruction.}
When $\mathcal{G}^{\text{stage}}$ is being optimized, we can obtain the rendered depth map $D^{\text{render}}$ at frame $f$ by utilizing the Gaussian rasterization as follows:
\begin{equation}
D^{\text{render}} = \sum^{K}_{k=1}d_k\alpha_k\prod_{k'=1}^{k-1}(1-\alpha_{k'})
\label{eq:depth_rendering}
\end{equation}
where $d_k$ denotes the z-depth, and $\alpha_k$ is the blending coefficient of the \textit{k}-th Gaussian in view space.

Given $D^{\text{render}}$, we incorporate depth guidance in the log-L1 form~\cite{turkulainen2024dn} for better convergence of 3DGS as follows:
\begin{equation}
\mathcal{L}_{\text{depth}} = {\frac{\log\big(1+|M^{\text{stage}}D^{\text{render}}-M^{\text{stage}}D^{\text{aligned}}|\big)}{|M^{\text{stage}}|}},
\label{eq:depth_loss}
\end{equation}
where $M^{stage}$ is background mask obtained from Sec.~\ref{sec:preprocessing}.  We also add total variation loss for fostering the smoothness \cite{turkulainen2024dn, chung2024depth} of rendered depth $D^{\text{render}}$ as follows:
\begin{equation}
\mathcal{L}_{\text{TV}} = \frac{1}{|D^{\text{render}}|}\sum_{\mathbf{u}}
\Bigl|\frac{\partial D^{\text{render}}}{\partial \mathbf{u}}\Bigl|_1
\label{eq:TV_loss}
\end{equation}

As a result, our extended loss in addition to vanilla 3DGS losses ($\mathcal{L}_{color}$ and $\mathcal{L}_{\text{D-SSIM}}$) is defined as follows: 
\begin{equation}
\begin{split}
\mathcal{L}_{\text{background}} = & \ (1 - \lambda_{\text{D-SSIM}}) \mathcal{L}_{\text{color}} + \lambda_{\text{D-SSIM}} \mathcal{L}_{\text{D-SSIM}} \\
& + \lambda_{\text{depth}} \mathcal{L}_{\text{depth}} + \lambda_{\text{smooth}} \mathcal{L}_{\text{TV}},
\end{split}
\label{eq:background_loss}
\end{equation}
where we empirically set $\lambda_{\text{D-SSIM}}\!=\!0.2$, $\lambda_{\text{depth}} \!=\! 0.2$ and $\lambda_{\text{smooth}} \!=\! 0.5$. We mask actors appearing in depth maps and input images using $M^\text{stage}$ for stability when optimizing Eq.~(\ref{eq:background_loss}). An example of $\mathcal{G}^{\text{stage}}$ is shown in Fig.~\ref{fig:pipeline}.

\vspace{2mm}\noindent\textbf{Object removal.}
\begin{figure}[t]
    \centering
    \includegraphics[width=\linewidth]{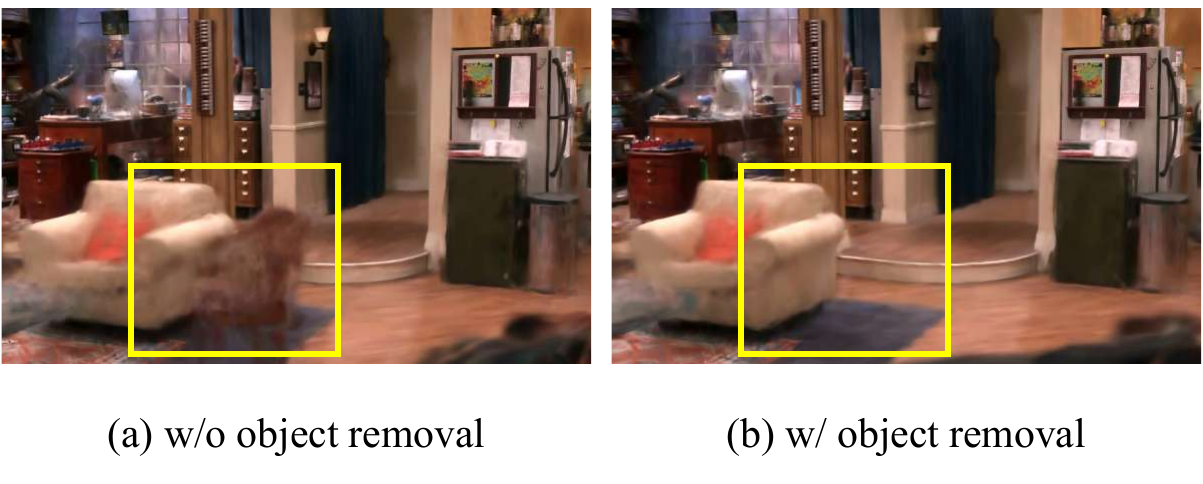}
    \vspace{-8mm}
    \caption{An example of transient \textbf{object removal}.}
    \label{fig:obj_remove}
    \vspace{-2mm}
\end{figure}
Since the gathered images have transient objects in the scene, they interfere with the reconstruction process. In particular, objects with no reference frames are hard to reconstruct or remove, which leads to floaters remaining in the background. These artifacts degrade the background quality by a large margin. To mitigate this issue, we annotate these regions and apply image inpainting~\cite{zhao2025ObjectClear}. Since inpainted areas can be noisy, we apply only the depth loss for robustness. We can successfully recover these regions, as shown in Fig.~\ref{fig:obj_remove}. 


\subsection{Locating Actors on the Stage}\label{sec:positioning}
Estimating human poses from a 2D image is a well-developed problem. We estimate the shape parameter $\beta \in \mathbb{R}^{10}$ and pose parameter $\theta \in \mathbb{R}^{24 \times 3 \times 3}$ of a SMPL model using the off-the-shelf approach~\cite{newell2025comotion}. However, when it comes to locating the humans in the designated 3D coordinate system, it becomes a nontrivial problem. 

Previous approaches proposed optimizing the human scale and pose with two adjacent shots~\cite{pavlakos2022one} or identifying the foot intersection point between a SMPL model and the ground plane~\cite{guo2023vid2avatar, jiang2022neuman,kocabas2024hugs}. However, these methods are inadequate for TV show scenarios since consecutive shots do not always feature the same individuals, and actors frequently have their feet out of the frame.

\vspace{2mm}\noindent\textbf{Spatio-temporal positioning module.}
We propose a \textit{spatio-temporal positioning} module that positions the posed humans to the reconstructed 3D stage using a single clip. 
Specifically, given \textit{i}-th SMPL vertex $\textbf{c}_i$ in canonical space, we can apply an arbitrary human pose to it by applying $\textbf{R}_{\theta}$, a linear blending skinning transformation induced by pose $\theta$:
\begin{equation}
\textbf{v}_i = \textbf{R}_{\theta}(\textbf{c}_i) = \sum_{j=1}^J w_{i,j}(\textbf{R}_{\theta,j} \textbf{c}_{i} + \textbf{t}_{\theta,j}),
\label{eq:LBS}
\end{equation}
where $w_{i,j} \in \mathbb{R}$ is the LBS weights of the $j$-th joint and $i$-th vertex, and \{$\textbf{R}_{\theta,j}$, $\textbf{t}_{\theta,j}$\} are the rotation and translation of $j$-th joint determined by predicted SMPL pose parameter $\theta$ with $J$ being the total number of joints. Then, posed SMPL vertices $\mathbf{v}_i$ can be mapped to the stage coordinate as follows:
\begin{equation}
    \mathbf{v}_i'(s, \textbf{t}) = s \textbf{v}_i + \textbf{t}.
\label{eq:vertex_alignment}
\end{equation}
positioning module finds the optimal scale $s\in \mathbb{R}$ and global translation parameters $\textbf{t}\in \mathbb{R}^3$ of the posed SMPL.

\vspace{2mm}\noindent\textbf{Depth loss.}
The core idea of our \textit{spatio-temporal positioning} module is to align posed SMPL vertices $v_i'$ to the aligned mono-depth $D^{\text{aligned}}_f$ (Sec.~\ref{sec:preprocessing}). Note that this scheme does not assume the same actor between consecutive shots~\cite{pavlakos2022one}, and does not involve the ground plane assumption~\cite{jiang2022neuman,kocabas2024hugs}.

For the alignment, we only consider SMPL's visible points $\mathbf{v}'\in\mathcal{V}_f$ at frame $f$ since the human regions of the $D^{\text{aligned}}_f$ indicate the depth of the visible parts.
We use Huber loss from Eq.~\eqref{eq:huber_loss} between the z-value of the frontal vertices in the camera space and $D^{\text{aligned}}_f$.
\begin{equation}
\mathcal{L}(\textbf{v}', \delta_2;s,\textbf{t}) = \textrm{Huber}(D^{\text{aligned}}(v'_x, v'_y), v'_z)
\label{eq:huber_loss2}
\end{equation}
where we empirically set $\delta_2=r^{\text{stage}}/20$.
We use the optimal $s^*$ and $\textbf{t}^*$ to place SMPL model into frame $f$.

\vspace{2mm}\noindent\textbf{2D Keypoint loss.}
Maintaining consistency between projected SMPLs and 2D image is important because misalignment degrades the quality of trained actor gaussians.
To keep SMPLs aligned to the image, we first estimate 2D keypoints by utilizing off-the-shelf 2D human pose detection module~\cite{yang2023effective}. Then we compare them with the projected joints of SMPLs to compute a keypoint loss $\mathcal{L}_{kpt}$ and optimize the SMPL pose parameter $\theta$. We use a robust Geman-McClure function~\cite{geman1987statistical} as follow:
\begin{equation}
\rho_{GM}(\gamma;\tau) = \frac{\gamma^2}{\gamma^2+\tau^2}
\label{eq:GMC}
\end{equation}
\begin{equation}
\mathcal{L}_{kpt}(\theta, \tau; s, \textbf{t}) = \sum_{j=1}^{J}c_j{\rho}_{GM}(||\mathit{\pi}_f(\mathcal{J}'_j)-\hat{\mathcal{J}}^{\mathrm{2D}}_{j}||_1;\tau)
\label{eq:kpt}
\end{equation}
where $\mathcal{J}'_j = sR_\theta(\mathcal{J}(\beta)_j) + \textbf{t}$, represents the 3D position of the j-th skeleton joint derived from the SMPL shape parameter $\beta$, global scale and translation $\{s, \textbf{t}\}$. Here, $\{\mathcal{J}(\cdot)_j\}$ denotes the joint positions of a SMPL model in the canonical space, $\{\hat{\mathcal{J}}^{\textrm{2D}}_{j}\}$ are the estimated 2D human keypoints, and $c_j$ represents the confidence score for the j-th keypoint detection. We empirically set $\tau$ as 1,000.

\begin{figure}[t]
    \centering
    \includegraphics[width=0.5\textwidth]{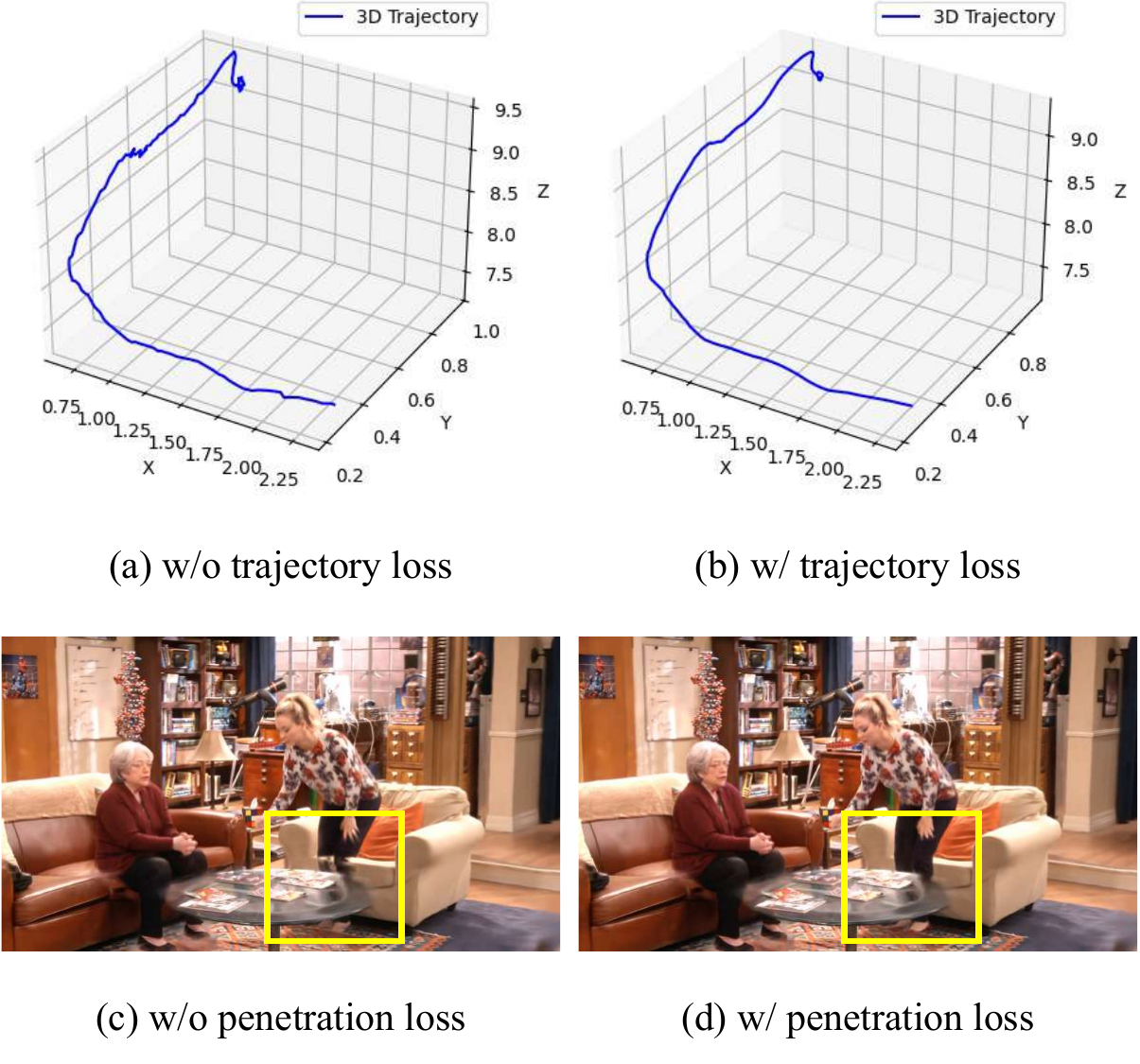}
    \vspace{-8mm}
    \caption{
    Effects of trajectory loss and penetration loss in \textit{spatio-temporal positioning} module.
    }
    \vspace{-2mm}
    \label{fig:positioning}
\end{figure}

\vspace{2mm}\noindent\textbf{3D Trajectory loss.}
Even when SMPLs are accurately aligned to each 2D image, their motion can still be unnatural in global space, due to depth inconsistencies across frames. 
To ensure smooth actor motion in 3D space while preserving 2D keypoint alignment, we adopt jerk loss~\cite{flash1985coordination}, which suppresses jittering movement of the actors. 
Jerk loss is defined as the squared difference between successive third-order finite differences of positions.
\begin{align}
\mathcal{L}_{traj}(\boldsymbol{\theta}; s, \mathbf{t})
&=
\frac{1}{(F-3)\,J}
\sum_{f=4}^{F}\sum_{j=1}^{J}
\bigl\|\,\Delta^{3}\mathcal{J}'_{f,j}\,\bigr\|_{2}^{2},\\
\label{eq:traj}
\qquad
\Delta^{3}\mathcal{J}'_{f,j}
&=
\mathcal{J}'_{f,j}-3\mathcal{J}'_{f-1,j}+3\mathcal{J}'_{f-2,j}-\mathcal{J}'_{f-3,j},
\end{align}
Here, $\mathcal{J}'_{f,j}$ represents the 3D position of the j-th skeleton joint at frame index $f$, and $F$ is the total number of frames. We denote $\boldsymbol{\theta}$ as $\{\theta_f\}_{f=1..F}$. As shown in Fig~\ref{fig:positioning}, $\mathcal{L}_{traj}$ produces a smoother 3D trajectory.

\vspace{2mm}\noindent\textbf{Penetration loss.}
Compositional video reconstruction of both actors and the stage requires their seamless integration.
To prevent actors from penetrating into the 3D stage, we use penetration loss as follows:
\begin{equation}
\mathcal{L}_{penet}(\mathcal{V}'_f;s, \textbf{t})
= \frac{1}{|\mathcal{V}'_f|}\sum_{\textbf{v}' \in \mathcal{V}'_f}[v'_{z} - D^{render}(v'_{x},v'_{y})]_+
\label{eq:penet}
\end{equation}
$D^{render}$ is a rendered depth map from the 3D stage (Eq.~(\ref{eq:depth_rendering})) and $[x]_+$ is $max(0,x)$. $\mathcal{V}'_f$ are the visible vertices of posed SMPL in image space at frame $f$. 
We penalize if the depth of the SMPL vertices exceeds the corresponding stage depth map. 
Since we select vertices that are not occluded by the foreground, penetration loss can handle scenarios where part of the actor is positioned behind stage objects as in Fig.~\ref{fig:positioning}.

\vspace{2mm}\noindent\textbf{Total loss.}
Total loss for positioning the actors can be summarized as follows. We iterate the alignment process for every actor.
\begin{equation}
\begin{split}
\mathcal{L}_{pos} = \
\lambda_{depth}\mathcal{L}_{depth}
+ \lambda_{kpt}\mathcal{L}_{kpt} \\
+ \lambda_{traj}\mathcal{L}_{traj}
+ \lambda_{penet}\mathcal{L}_{penet}
\end{split}
\label{eq:align_loss}
\end{equation}
We empirically set $\lambda_{\text{depth}}\!=\!1.0$, $\lambda_{\text{kpt}}\!=\!1.0$, $\lambda_{\text{traj}}\!=\!0.5$, and $\lambda_{\text{penet}}\!=\!0.001$. We use the Adam optimizer and the exponential learning rate scheduler for optimization.

\subsection{Tracking Actors across the Shots}\label{sec:tracking}
\textit{Spatio-temporal positioning} module (Sec.~\ref{sec:positioning}) places SMPL actors on the stage for every frame. Using the modern approach~\cite{newell2025comotion}, the SMPLs are associated across frames. However, it is still necessary to associate the SMPL model with the different shots to avoid creating multiple actors. To address this issue, we propose a ShotMatcher module to associate actors between shot boundaries.

\begin{figure}[t]
    \centering
    \includegraphics[width=\linewidth]{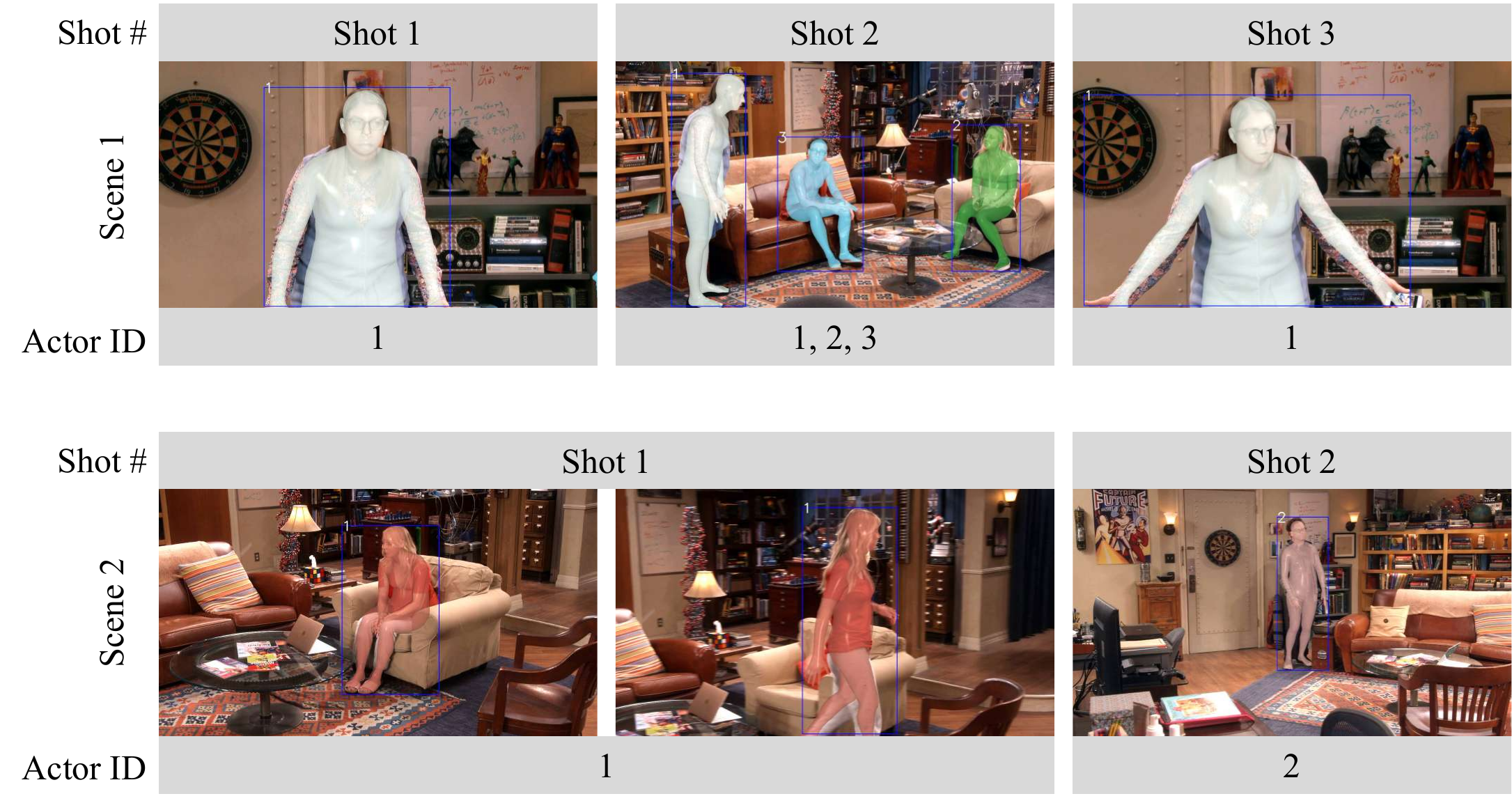}
    \vspace{-6mm}
    \caption{Results of \textbf{actor association}. \textit{ShotMatcher} can associate actors even when some individuals do not appear in a shot. If the distance of the matched actors is above the matching threshold, \textit{ShotMatcher} identifies them as different.}
    \label{fig:actor_association}
    \vspace{-2mm}
\end{figure}

As shown in Fig.~\ref{fig:actor_association}, ShotMatcher calculates the pairwise Euclidean distances between the actors' 3D coordinates in the last frame of a certain shot and the first frame of the consecutive shot. Among all possible actor-to-actor pairs, ShotMatcher chooses the actor pair with the smallest Euclidean distance that falls below a matching threshold. As shown in Fig.~\ref{fig:actor_association}, the matching threshold is required to exclude the pair with a far distance. Details are described in Algorithm~\ref{alg:match}.

\begin{algorithm}[t]
\caption{Actor association algorithm}
\begin{algorithmic}[1]
\STATE \textbf{Input:}
\begin{itemize}
    \item $F_i$: last frame of the previous shot
    \item $F_{i+1}$: first frame of the subsequent shot
    \item $A = \{ A_1, A_2, \dots, A_n \}$: centers of $N$ actors in $F_i$
    \item $B = \{ B_1, B_2, \dots, B_m \}$: centers of $M$ actors in $F_{i+1}$
    \item $\lambda$: matching threshold
\end{itemize}
\STATE \textbf{Output:} $P$: matched pairs set
\STATE \textbf{Begin:}
\STATE $P$ $\gets$ $\emptyset$
\STATE $B^{\textit{unmatched}}$ $\gets$ $B$

\FOR{$A_i \in A$}
    \STATE $min\_distance \gets \infty$
    \STATE $B^{\textit{selected}} \gets$ \textbf{None}
    \FOR{$B_j \in B^{\textit{unmatched}}$}
        \STATE $d \gets \text{EuclideanDistance}(A_i,B_j)$
        \IF{$d < min\_distance$}
            \STATE $min\_distance \gets d$
            \STATE $B^{\textit{selected}} \gets B_j$
        \ENDIF
    \ENDFOR
    \IF{$min\_distance < \lambda$}
        \STATE $P \gets P \cup \{ (A_i, B^{\textit{selected}}) \}$
        \STATE $B^{\textit{unmatched}} \gets B^{\textit{unmatched}} - B^{\textit{selected}}$
    \ENDIF
\ENDFOR

\RETURN $P$

\end{algorithmic}
\label{alg:match}
\end{algorithm}

\vspace{2mm}
\noindent\textbf{Pose interpolation and extrapolation.}
Entertainment videos frequently present various occlusion scenarios, such as one actor blocking another, objects obscuring actors, or actors temporarily moving out of the frame. These occlusions interfere with estimating accurate human pose. To address this issue, we analyze SMPL tracking results, identify two stable frames adjacent to the occluded frames, and perform linear interpolation of the actors' SMPL parameters. 

In entertainment videos like TV shows, one of the main challenges of actor association is the absence of certain actors in some shots. For example, while a wide shot may include all actors on stage, a close-up shot might only feature one or two individuals. To address this, any unmatched actors are filled into the subsequent shots by extrapolating their SMPL parameters from the data at the shot boundary.


\subsection{3D Actor Reconstruction}\label{sec:3D_actor}
\begin{figure}[t]
    \centering
    \includegraphics[width=0.5\textwidth]{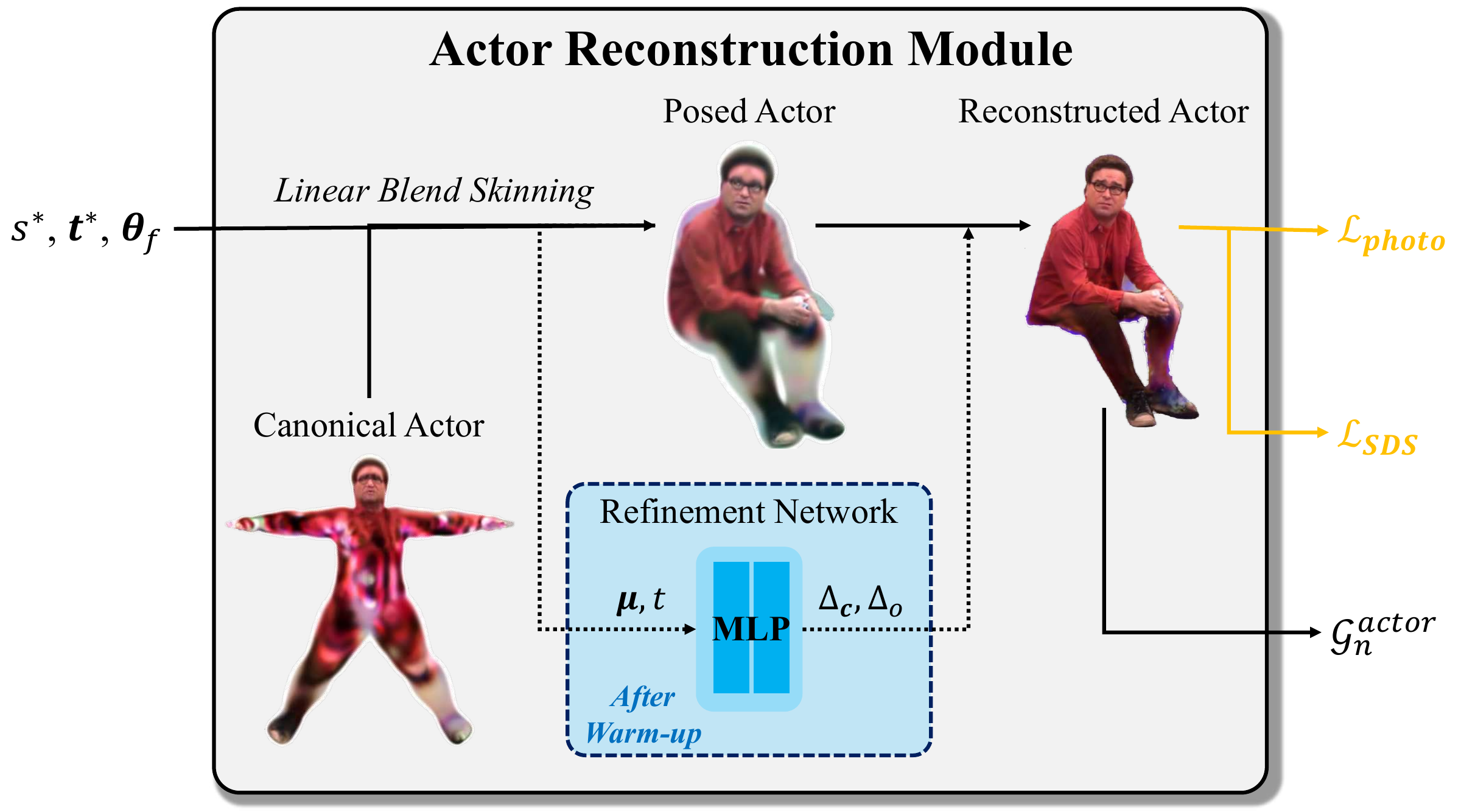}
    \vspace{-6mm}
    \caption{
    Overview of our actor reconstruction module. Actor gaussians are trained with photometric loss and SDS loss per actor. After warm-up, we jointly train refinement network to enhance details.
    }
    \label{fig:actor_module}
    \vspace{-2mm}
\end{figure}

In the final step of our pipeline, we introduce our human reconstruction module that makes $\{\mathcal{G}^{\text{actor}}_n\}_{n=1...N}$ using 3DGS, given $N$ SMPL models associated with different shots, as shown in Fig.~\ref{fig:actor_module}.
We first initialize Gaussian centers of $\mathcal{G}^{\text{actor}}_n$ using each of the SMPL model's vertices located in the stage coordinate (Sec.~\ref{sec:tracking}). Then, we optimize $\mathcal{G}^{\text{actor}}_n$ using the 3DGS loss extended with SDS loss $\mathcal{L}_{\text{SDS}}$~\cite{lee2024guess}. 

\vspace{2mm}
\noindent\textbf{Photometric loss.}
Since the stage can occlude the actors, we estimate foreground masks from the stage structure by comparing the depth between the rasterized background Gaussians $D^\text{render}_\text{stage}$ and the rasterized human Gaussians $D^\text{render}_\text{actor}$ using Eq.~(\ref{eq:depth_rendering}). If a rendered depth pixel of the actor is smaller than the stage, such pixel is marked as a part of the foreground as follows:
\begin{equation}
M^{\text{foregd}}(\mathbf{u}) = 
\begin{cases} 
1, & \!\!\text{if }  D^{\text{render}}_{\text{stage}}(\mathbf{u}) \!>\! \big\{D^{\text{render}}_{\text{actor}, n}(\mathbf{u})\big\}_{n=1...N} \\
0, & \!\!\text{otherwise},
\end{cases}
\label{eq:fore_mask}
\end{equation}
where $N$ denotes the number of humans.

\begin{figure}[t]
    \centering
    \includegraphics[width=0.8\linewidth]{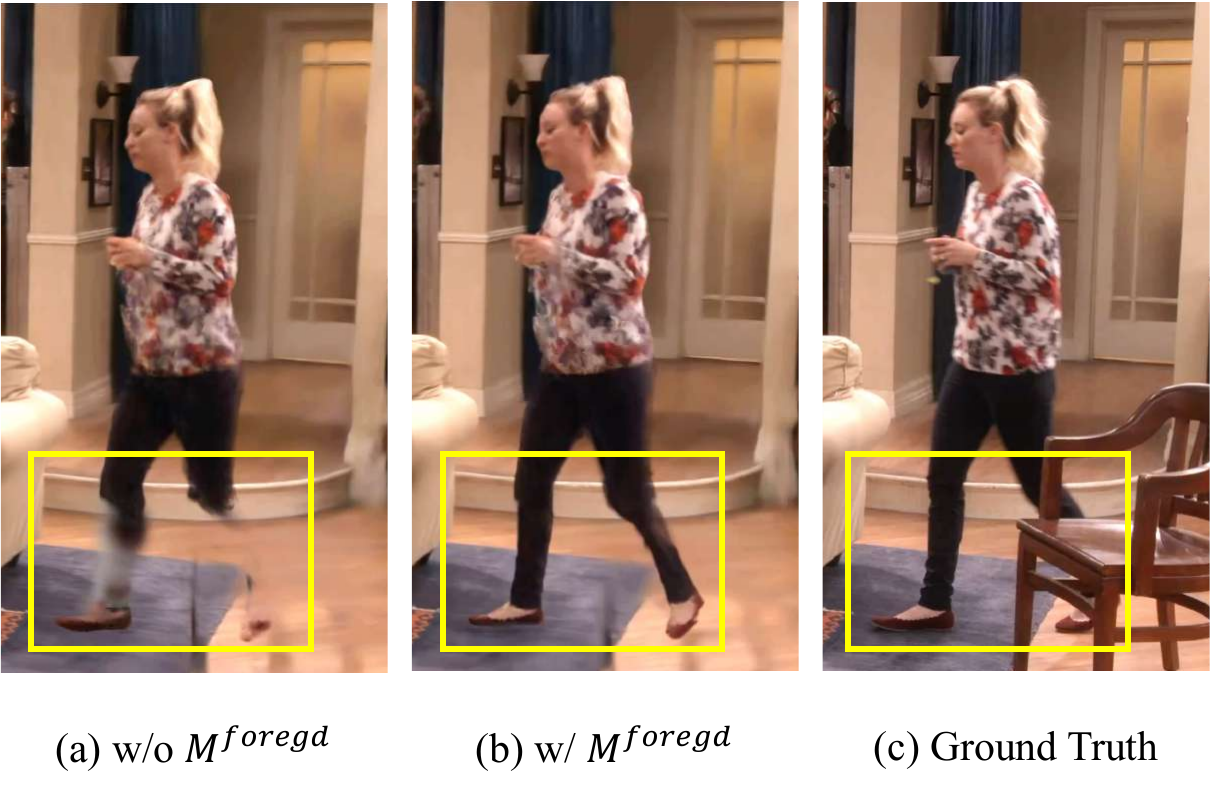}
    \vspace{-4mm}
    \caption{An effect of using the proposed \textbf{foreground masking}.}
    \label{fig:fore_mask}
    \vspace{-2mm}
\end{figure}

We then apply the foreground masks to both the rendered actors $P\!=\!M^{\text{foregd}}\odot I_{\text{actor}}^\text{render}$ and input video frame $Q\!=\!M^{\text{foregd}}\!\odot\! I$, and calculate actor loss $\mathcal{L}_{\text{actor}}$ using the two masked images $\{P,Q\}$ for the vanila 3DGS loss~\cite{kerbl20233d}, where $\mathcal{L}_{\text{actor}}$ is calculated for every visible frame of $n$-th actor. 
The effect of using a foreground mask is shown in Fig.~\ref{fig:fore_mask}.

\vspace{2mm}\noindent\textbf{Unobserved area.}
In entertainment videos like TV shows, cameras usually capture the actors from the front, leaving the rest of the part unobserved during the entire video clip. Inspired by \cite{lee2024guess}, we use SDS loss~\cite{poole2022dreamfusion} with textual inversion to hallucinate the unseen parts of the actors. We compute SDS loss for each Gaussian sets $\{\mathcal{G}^{\text{actor}}_n\}_{n=1...N}$ with person-specific diffusion $\phi_n$.
The final loss of 3DGS for an individual actor is defined as follows:
\begin{equation}
\mathcal{L}_{\text{total}} = \lambda_{\text{actor}}\sum_f^F\mathcal{L}_{\text{actor},f} + \lambda_{\text{SDS}}\mathcal{L}_{\text{SDS}}(\mathcal{G}^{\text{actor}}_n; \phi_n)
\label{eq:total_actor_loss}
\end{equation}

\vspace{2mm}
\noindent\textbf{Refinement.}
In entertainment videos like TV shows, recovering detailed facial expressions are essential for delivering emotion and enhancing realism. We introduce an implicit function-based residual appearance fitting scheme using a Gaussian deformation network~\cite{jung2023deformable}. In our scenario, we observe that changing the positions of Gaussians on the face is unsuitable since facial motions are subtle. Instead of moving the Gaussians, we refine colors and opacity to fit the details. Given time $t$ and the position $\boldsymbol{\mu}$ of Gaussians, the color and opacity head of the face-fitting network return each residual value as follows,
\begin{align} 
\Delta \mathbf{c} (\boldsymbol{\mu}, t) &= F_{\text{color}}\Big(F_{\theta}\big(\text{concat}\{\gamma(\boldsymbol{\mu}), \gamma(t)\}\big)\Big)\\
\Delta o (\boldsymbol{\mu}, t) &= F_{\text{opacity}}\Big(F_{\theta}\big(\text{concat}\{\gamma(\boldsymbol{\mu}), \gamma(t)\}\big)\Big)
\end{align}
where $\Delta \mathbf{c}$ and $\Delta o$ denotes color and opacity residual, \{$F_\theta$, $F_\text{color}$, $F_\text{opacity}$\} indicates learnable MLPs for deformation, and $\gamma(\cdot)$ is positional encoding. 

We first train actor Gaussians without the refinement module for 2,000 iterations to reconstruct coarse actors. Then, for the rest of the iterations, we add the color and the opacity residuals to the actor Gaussians obtained from Eq.~(\ref{eq:total_actor_loss}). For $n$-th actor, we can update it as follows:
\begin{equation}
\mathcal{G}^{actor}_{n,t} = \big\{g_{k,t}(\boldsymbol{\mu},\mathbf{s},\mathbf{q},\mathbf{c} + \Delta\textbf{c}, o + \Delta o)\big\}_{k=1...K}
\end{equation}
\begin{figure*}[t]
    \centering
    \includegraphics[width=0.8\textwidth]{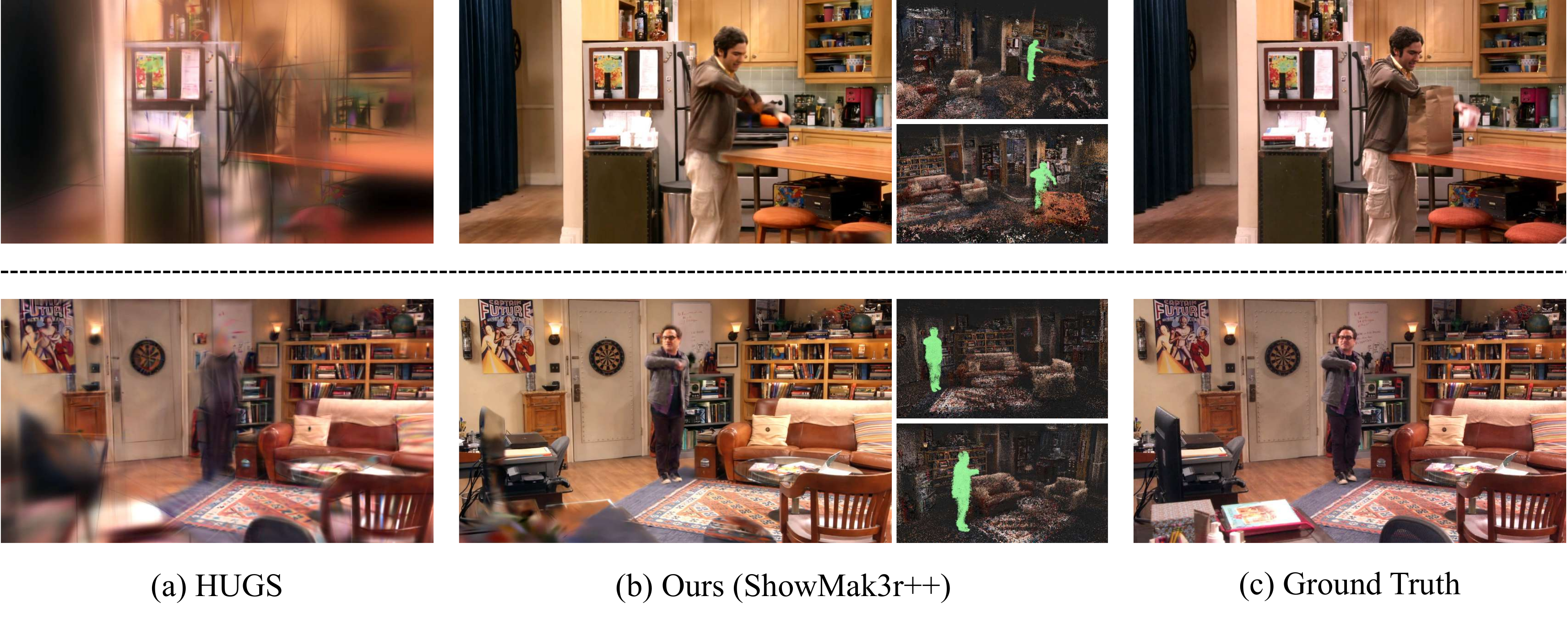}
    \vspace{-4mm}
    \caption{Qualitative comparison on 'The Big Bang Theory' videos from Sitcoms3D dataset, where each video feature a single actor. Our method demonstrates superiority in \textbf{accurately positioning actors} on the stage. Green points denote the Gaussian centers for the actors.}
    \label{fig:result_single}
\end{figure*}
\begin{figure*}[t]
    \centering
    \includegraphics[width=0.98\textwidth]{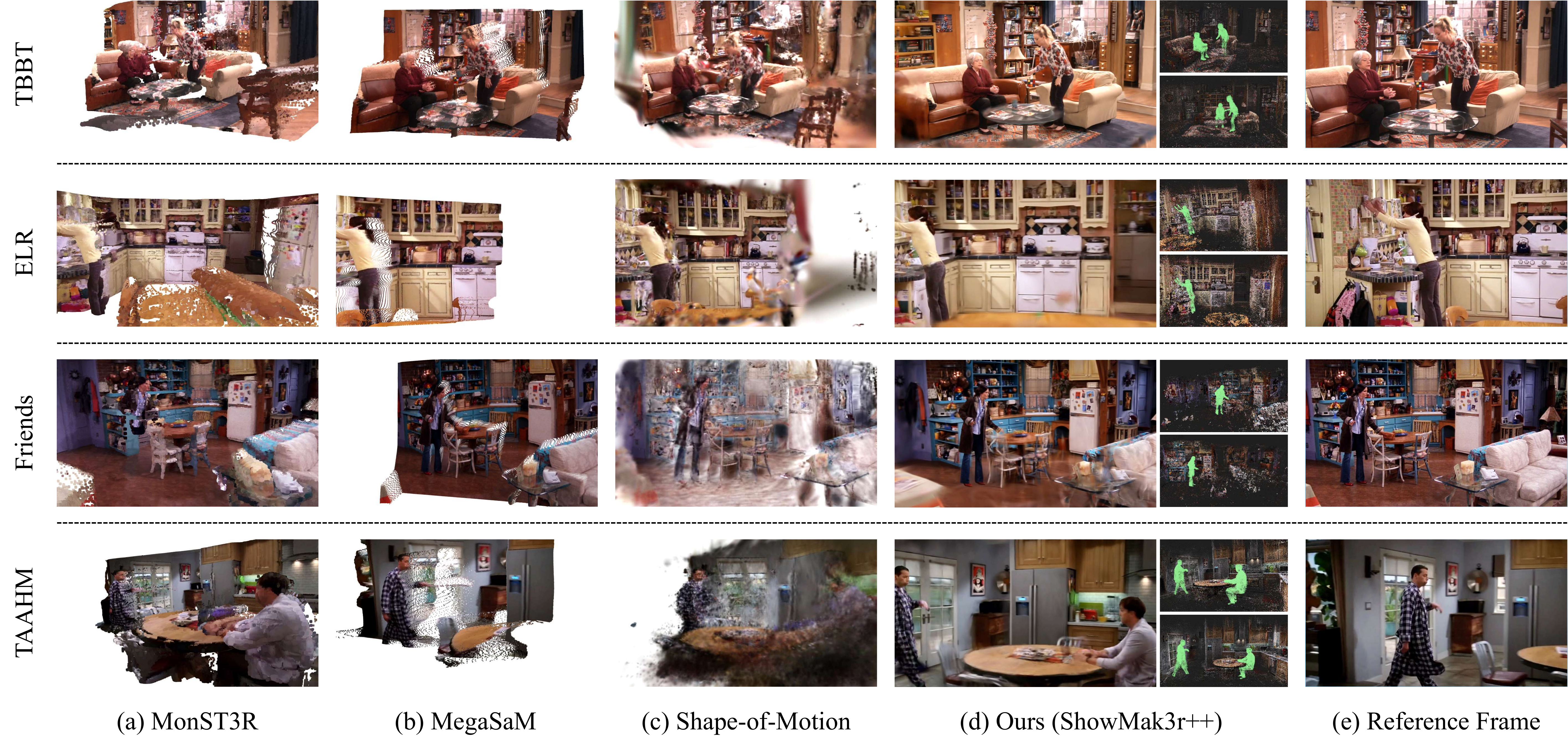}
    \vspace{-4mm}
    \caption{Qualitative comparison of four TV show videos, each featuring multiple actors. We compare our pipeline with both point cloud representation and Gaussian representation baselines. The frame from the input viewpoint is referred to as the 'Reference Frame'. Green points denote the Gaussian centers for the actors.}
    \label{fig:result_multi}
    \vspace{-2mm}
\end{figure*}
\begin{figure*}[t]
    \centering
    \includegraphics[width=0.8\textwidth]{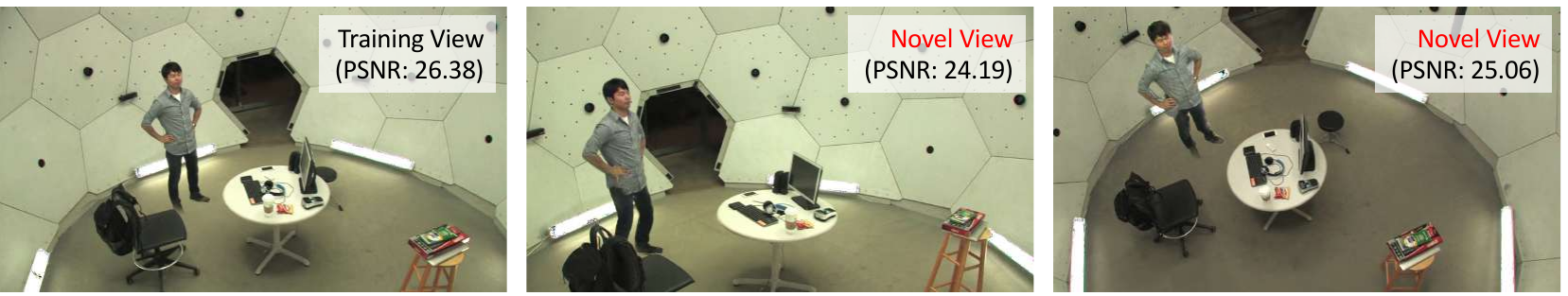}
    \vspace{-2mm}
    \caption{Qualitative result on CMU Panoptic dataset. Not only does our method produce photometric results from novel viewpoints, but it also aligns the human to the correct position. }
    \label{fig:cmu}
\end{figure*}
\begin{figure*}[t]
    \centering
    \includegraphics[width=0.98\textwidth]{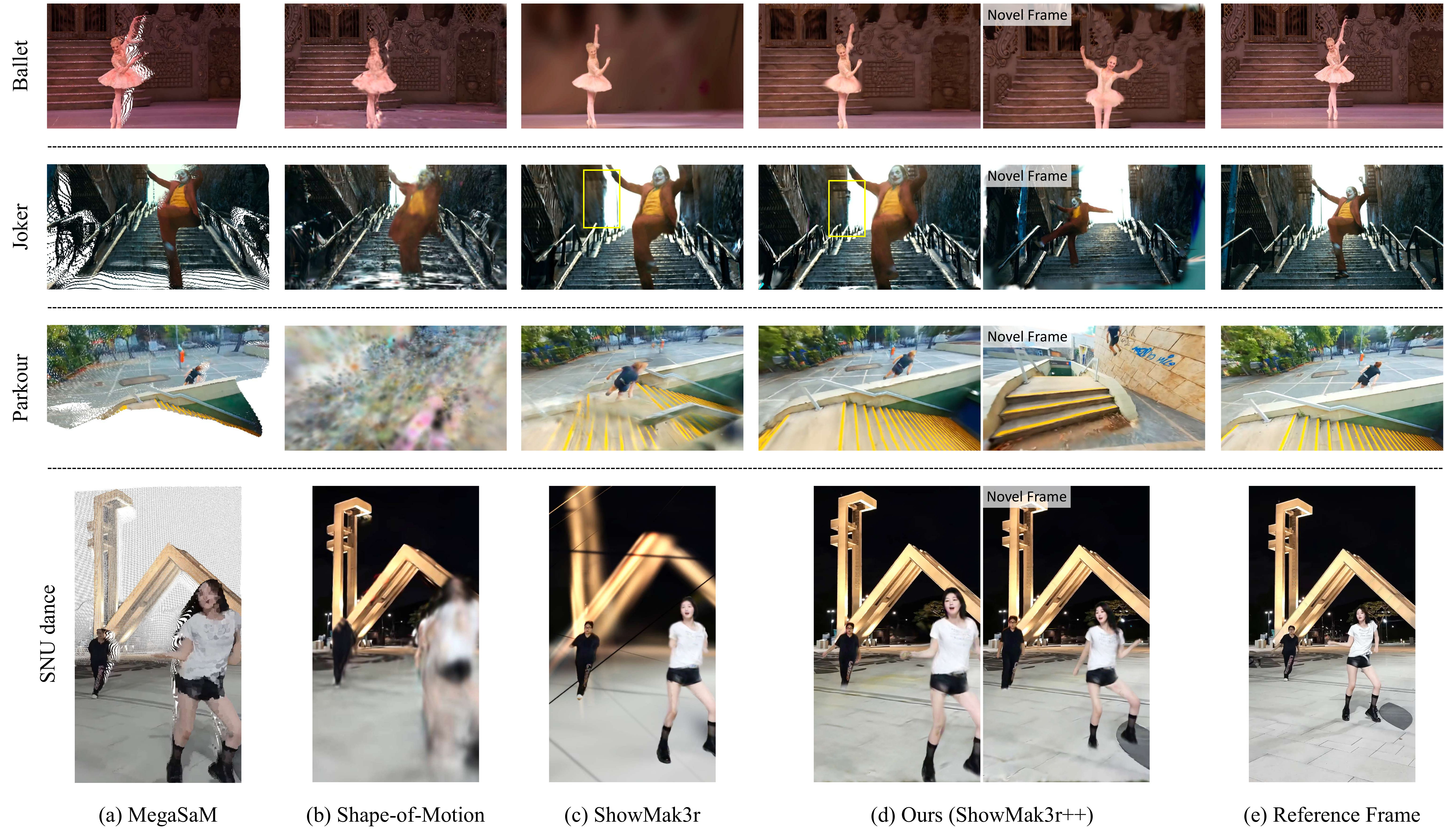}
    \vspace{-4mm}
    \caption{
    Qualitative comparison of web videos. Our method can handle various scenarios such as dynamic action clips, dance videos, or movie clips. The left side of (c) shows results from a novel viewpoint, and the right side shows results from a novel frame.}
    \vspace{-2mm}
    \label{fig:result_web}
\end{figure*}

\section{Experiments}
\subsection{Experiment Setups}
\label{sec:setups}
We evaluate our pipeline on Sitcoms3D and CMU Panoptic dataset~\cite{pavlakos2022one, joo2015panoptic}. Sitcoms3D dataset consists of 100-200 images per location from several sitcoms, such as '\textit{The Big Bang Theory} (2007)', '\textit{Friends} (1994)', '\textit{Two and a Half Men} (2003)', '\textit{Everybody Loves Raymond} (1996)'. These images are sampled across multiple episodes to capture the total structure of the stage. 
We also evaluate our pipeline in uncontrolled settings. We select challenging scenarios from web videos such as 'Ballet', 'Joker', 'Parkour', 'SNU dance'.
For qualitative comparison, we compare our pipeline with three different types of dynamic scene reconstruction approaches on four sitcoms of the Sitcoms3D dataset~\cite{pavlakos2022one}. We select HUGS~\cite{kocabas2024hugs}, Shape-of-Motion~\cite{som2024}, 
MonST3R~\cite{zhang2024monst3r}, and MegaSaM~\cite{li2025megasam}
as baselines, each representing template-based, template-free, and last two as feed-forward methods. 
We also compare the stage reconstruction quality with other 3D reconstruction methods such as Sitcoms3D(NeRF-W~\cite{martin2021nerf}), 3DGS~\cite{kerbl20233d}, GS-W~\cite{zhang2024gaussian}, and FSGS~\cite{zhu2024fsgs}. We calculate PSNR, SSIM, and LPIPS metrics with masked human areas excluded for a fair comparison. 
Lastly, to evaluate the effectiveness of our face-fitting network, we compare our method with ExAvatar~\cite{moon2024expressive}, which also addresses similar challenges of facial expression reconstruction.

\subsection{Qualitative Comparison}
\label{sec:qualitative_comparison}
We compare our method with HUGS~\cite{kocabas2024hugs} (template-based), Shape-of-Motion~\cite{som2024} (template-free), MonST3R~\cite{zhang2024monst3r} and MegaSaM~\cite{li2025megasam} (feed-forward). We compare the viewpoints that depict the stage for a fair comparison, since other baselines target reconstruction from a monocular video. 

\vspace{2mm}\noindent\textbf{Single person scenario.}
We compare HUGS, which targets single-person scenarios, with the videos featuring a single individual. HUGS utilizes an SfM system for camera pose estimation, so it can not reconstruct the scene from a monocular video clip without sufficient view changes. Therefore, we provide additional background images to HUGS pipeline for camera pose prediction. Although our approach also utilizes SfM for the pose estimation, our depth guidance-based approach helps reconstruct the 3DGS scenes even with minimal camera movement. As shown in Fig.~\ref{fig:result_single}, HUGS struggles to accurately locate actors in TV show videos, whereas our pipeline effectively reconstructs both actors and stages. Note that HUGS assumes the foot is visible and touches the ground, whereas our pipeline does not.

\vspace{2mm}\noindent\textbf{Multiple people scenario.}
Shape-of-Motion~\cite{som2024}, MonST3R~\cite{zhang2024monst3r}, and MegaSaM~\cite{li2025megasam} do not rely on template-based models, so we compare them with more challenging multi-people scenarios. We select video clips with a single shot since they assume continuous viewpoints.
As demonstrated in Fig.~\ref{fig:result_multi}, both Shape-of-Motion and MonST3R face challenges in deforming dynamic actors in a moving camera, even when videos with a single shot are given. 
MegaSaM predicts better results than MonST3R, but it predicts in point cloud representation, which is not photo-realistic. Additionally, all baselines only consider monocular video settings, making it challenging to utilize additional background images.
Our method, however, shows robust reconstruction from videos with arbitrary camera translations. 

\vspace{2mm}\noindent\textbf{CMU Panoptic dataset result.}
We evaluate additional qualitative results from CMU Panoptic dataset~\cite{joo2015panoptic}. This dataset captures multiple people interacting with each other within the multi-view camera system. To simulate a TV show within the dataset, we select 8 cameras (out of the original 31) that capture frontal views of the human subject. These 8 views are used for stage reconstruction, while only one among them is used for actor reconstruction. Fig.~\ref{fig:cmu} illustrates novel view synthesis results, achieving a PSNR of 25.21 on average. 

\vspace{2mm}\noindent\textbf{Web video result.}
As shown in Fig.~\ref{fig:result_web}, our method can handle extreme camera movements, challenging human poses, and outdoor scenarios, whereas state-of-the-art reconstruction models like Shape-of-Motion fail. 
We also compare our method with MegaSaM, which utilizes a different representation. Since the output is in point cloud format, the results are not photo-realistic. Also, MegaSaM faces difficulty predicting dynamic human poses because it does not utilize any human priors.
Additionally, we observe that even ShowMak3r~\cite{kim2025showmak3r} fails under challenging web video scenarios.

\subsection{Quantitative Comparison}
\label{sec:quantitative_comparison}
\begin{table}[t]
\centering
\caption{Quantitative comparison of \textbf{stage reconstruction results}: We present the reconstruction metrics for the living room from 'The Big Bang Theory' in the Sitcoms3D dataset.}
\resizebox{0.8\columnwidth}{!}{
    \begin{tabular}{lcccc}
        \toprule
        Methods\hspace{7mm} & PSNR$\uparrow$ & SSIM$\uparrow$ & LPIPS$\downarrow$ \\
        \midrule
        Sitcoms3D~\cite{pavlakos2022one} & 18.81 & 0.62 & 0.55 \\
        3DGS~\cite{kerbl20233d} & 19.21 & 0.64 & 0.49 \\
        GS-W~\cite{zhang2024gaussian} & 19.35 & 0.65 & 0.51 \\
        FSGS~\cite{zhu2024fsgs} & 19.34 & 0.65 & 0.49 \\
        \midrule
        Ours & \textbf{19.65} & \textbf{0.66} & \textbf{0.49} \\
        \bottomrule
    \end{tabular}
}
\label{tab:stage_table}
\vspace{-5mm}
\end{table}

\noindent\textbf{Stage reconstruction.} We compare our stage reconstruction quality with other 3D reconstruction methods. For a fair comparison, we do not use object removal (Sec.~\ref{sec:3D_stage}). Given 165 background images of the TBBT scene in the Sitcoms3D dataset, we reconstruct the background except for 10 randomly picked images for the testing. As shown in Table~\ref{tab:stage_table}, GS-W~\cite{zhang2024gaussian} faces challenges removing transient actors. Also, compared to 3DGS~\cite{kerbl20233d}, and FSGS~\cite{zhu2024fsgs}, masking out actors and using depth priors improves the quality when reconstructing from aggregated sitcom images.

\begin{figure}[t]
\includegraphics[width=0.995\linewidth]{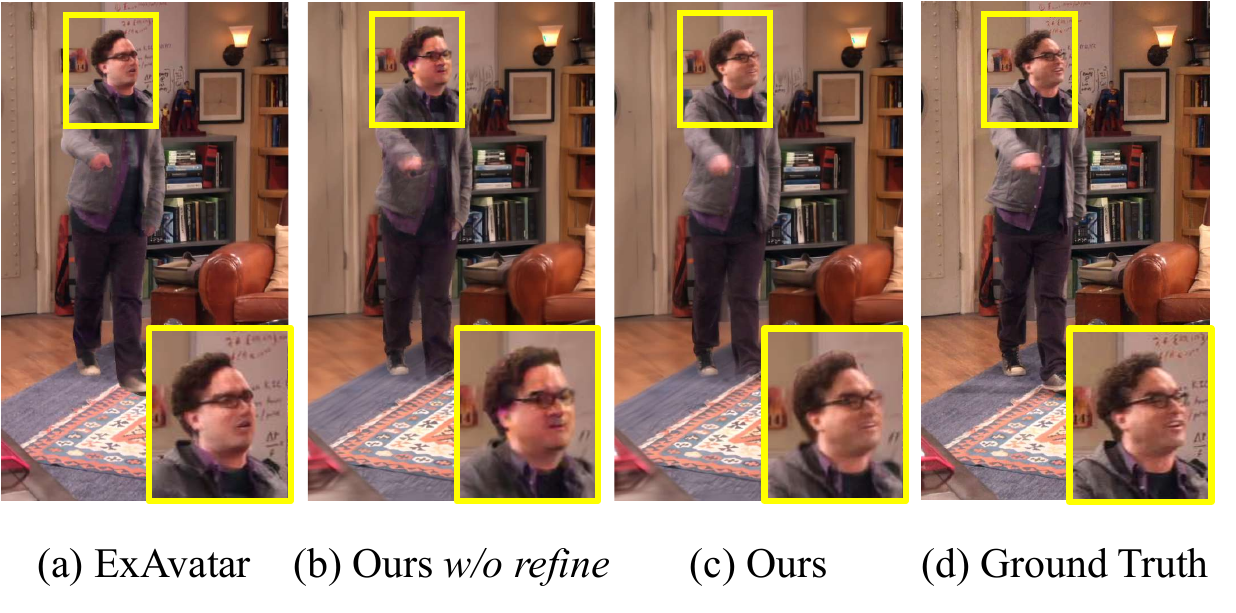}
\vspace{-8mm}
\caption{Ablation study for face-fitting network.}
\vspace{-2mm}
\label{fig:face-fitting}
\end{figure}

\vspace{2mm}\noindent\textbf{Actor reconstruction.}
We compare our face-fitting network with ExAvatar~\cite{moon2024expressive}. As shown in Fig.~\ref{fig:face-fitting}, while ExAvatar is designed to capture facial expressions, it struggles to recover fine details. In contrast, our simple and practical deformation module enables more precise facial expression adjustments.

\begin{table}[t]
\centering
\caption{Quantitative comparison of face reconstruction results. We compare the performance of our face-fitting network against ExAvatar~\cite{moon2024expressive}. For this comparison, we crop the facial region.}
\resizebox{0.85\columnwidth}{!}{
    \begin{tabular}{lcccc}
    \toprule
    Methods & PSNR$\uparrow$ & SSIM$\uparrow$ & LPIPS$\downarrow$\\ 
    \midrule
    ExAvatar~\cite{moon2024expressive} & 20.17 & 0.64 & \textbf{0.25} \\
    \midrule
    Ours \textit{w/o refinement} & 21.52 & 0.82 & 0.36 \\
    Ours & \textbf{24.34} & \textbf{0.84} & 0.28\\ \bottomrule
    \end{tabular}
}
\label{tab:face_table}
\end{table}

\subsection{Ablation Study}
\label{sec:ablation_study}
To evaluate the performance of \textit{spatio-temporal positioning} module, we propose two metrics: MTED (Mean Translation Euclidean Distance) and MPED (Mean Pose Euclidean Distance). These metrics calculate the Euclidean Distance of the global translation between two SMPL models and the Euclidean Distance of 3D joints at shot boundaries. We select the scene from TBBT where two subsequent shots have no overlapping background, e.g., the livingroom and the kitchen. As shown in Table~\ref{tab:ablation_table}, without \textit{spatio-temporal positioning} module, fitting fails to align with the correct positions, reducing correlation of the actor between adjacent shots.

To evaluate the effect of penetration loss, we compare the actor region of input frames with the rendered frames. We select the scene from `The Big Bang Theory' where multiple actors interact heavily with the stage, such as actors sitting on a couch. As shown in Table~\ref{tab:ablation_table2}, penetration loss produces better results by preventing actors from penetrating the stage.

\begin{table}[t]
\centering
\caption{Ablation study for spatio-temporal positioning module. We report Mean Translation Euclidean Distance (MTED) and Mean Pose Euclidean Distance (MPED) across 2 consecutive shots.}
\resizebox{0.9\columnwidth}{!}{
    \begin{tabular}{p{4cm}ccc}
    \toprule
    Methods & MTED$\downarrow$ & MPED$\downarrow$ \\ \midrule
    Ours \textit{w/o positioning module} & 1.99 & 0.24 \\
    Ours & \textbf{1.19} & \textbf{0.10} \\ \bottomrule
    \end{tabular}
}
\label{tab:ablation_table}
\end{table}
\begin{table}[t]
\centering
\caption{Ablation study for penetration loss. For comparison, we crop the actor regions.}
\resizebox{0.9\columnwidth}{!}{
    \begin{tabular}{p{2.9cm}ccc}
    \toprule
    Methods & PSNR$\uparrow$ & SSIM$\uparrow$ & LPIPS$\downarrow$\\ \midrule
    Ours \textit{w/o penetration loss} & 32.99 & 0.970 & 0.039 \\
    Ours  & \textbf{33.57} & \textbf{0.971} & \textbf{0.038} \\ \bottomrule
    \end{tabular}
}
\vspace{-2mm}
\label{tab:ablation_table2}
\end{table}

\subsection{Applications}
\label{sec:applications}
\begin{figure}[t]
    \centering
    \includegraphics[width=\linewidth]{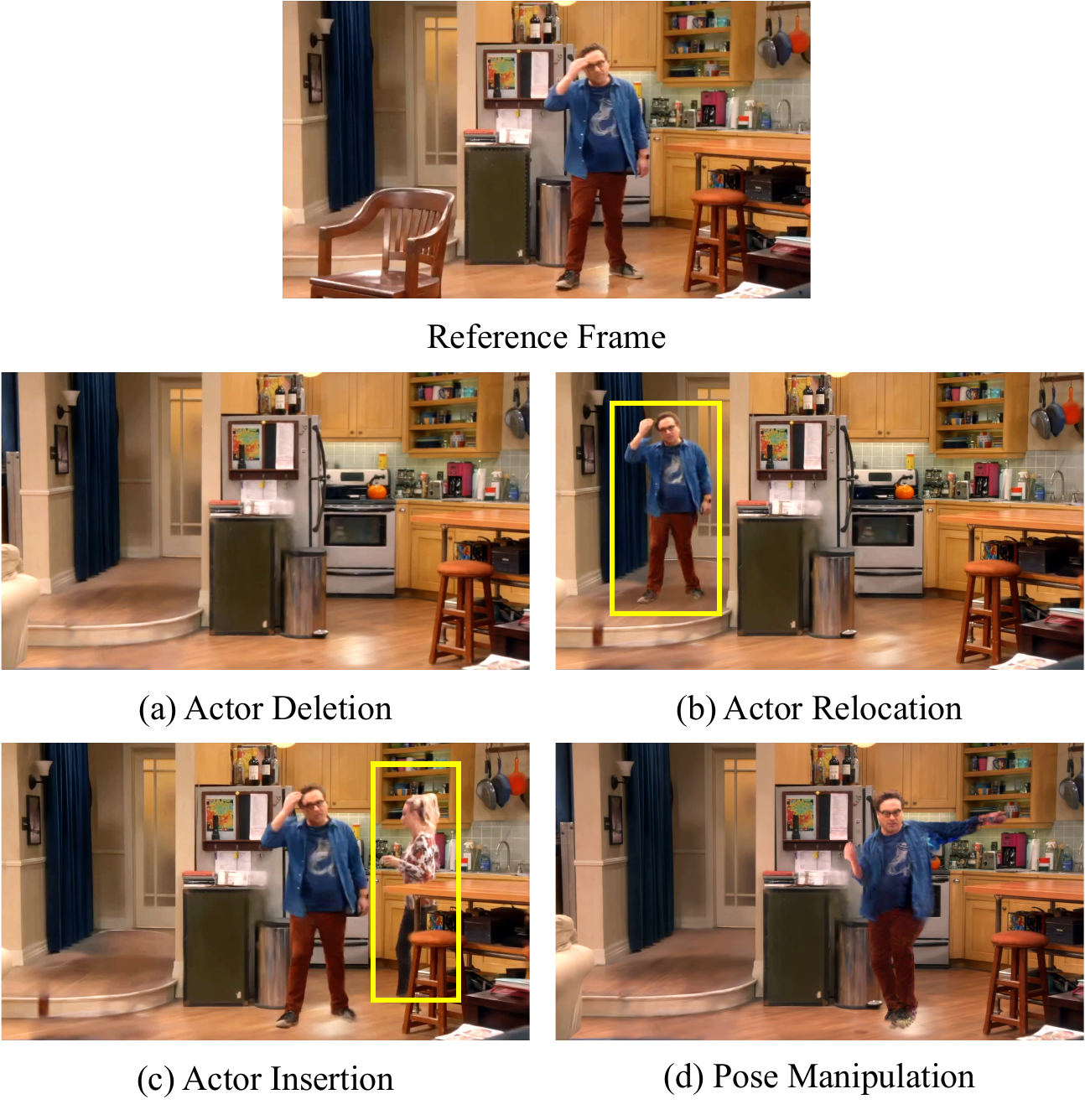}
    \vspace{-8mm}
    \caption{The reconstructed scenes with our pipeline are \textbf{editable}. Existing actors can be removed, repositioned, or replaced. New actors can be added, and their poses can be adjusted.}
    \label{fig:app}
    \vspace{-2mm}
\end{figure}
In Fig~\ref{fig:app}, we show the possible applications with our pipeline. Maintaining separate Gaussian sets for each actor and the stage makes it possible to perform edits, such as removing, relocating, and inserting specific actors from the video. Additionally, actor poses can be manipulated by controlling the pose parameters of the SMPL model.
\section{Conclusion}
We introduce a unified reconstruction pipeline that targets both controlled settings like TV shows and uncontrolled scenarios like web videos.
To tackle challenges in these environements, 
we propose the following key modules: (1) \textit{Spatio-temporal positioning} module, which positions actors on the stage by using depth prior and while maintaining 2D image alignment and natural 3D motions, (2) \textit{ShotMatcher}, which ensures continuous actor tracking across shot-changes, (3) face-fitting network for dynamic facial expression recovery. Experiments demonstrate that our method effectively reassembles entertainment videos from novel camera viewpoints. 

\noindent\textbf{Limitation and Future Work.}
Our current pipeline can only reconstruct the background of the web videos that have been observed during the video. We plan to address these limitations in future research.

\bibliographystyle{IEEEtran}
\bibliography{main}

\section{Biography}
\vspace{-12mm}
\begin{IEEEbiography}[{\includegraphics[width=1in,height=1in,clip,keepaspectratio]{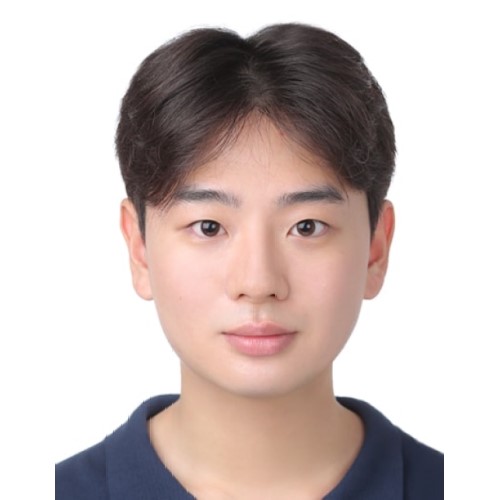}}]{Sangmin Kim}
is a Ph.D. student of the Interdisciplinary Program in Artificial Intelligence at Seoul National University, Republic of Korea. His research interests include 3D/4D computer vision applications for VFX and post-production.
\end{IEEEbiography}
\vspace{-12mm}
\begin{IEEEbiography}[{\includegraphics[width=1in,height=1in,clip,keepaspectratio]{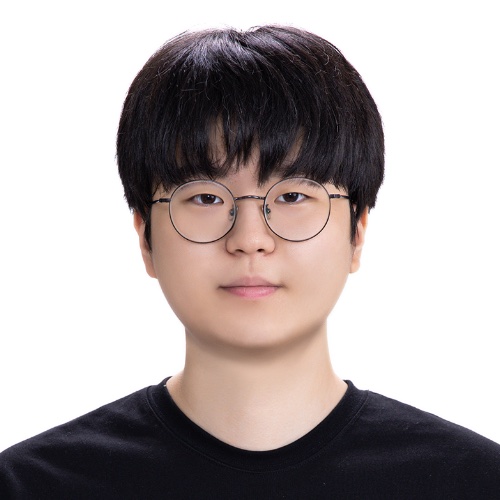}}]{Seunguk Do}
is a Ph.D. student of the Interdisciplinary program in Artificial Intelligence at Seoul National University, Republic of Korea. His research interests include 3D Hunan generative models and multi-modal representation learning,
especially in the 3D domain.
\end{IEEEbiography}
\vspace{-12mm}
\begin{IEEEbiography}[{\includegraphics[width=1in,height=1in,clip,keepaspectratio]{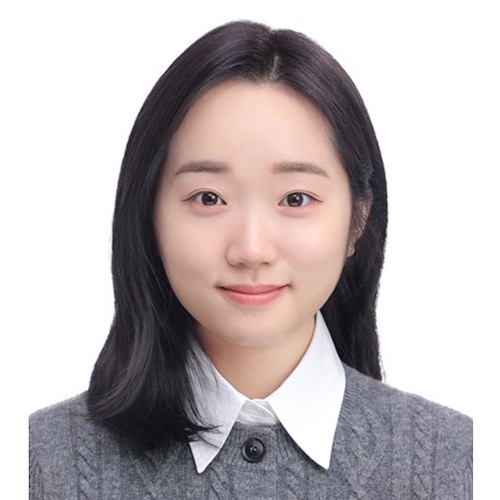}}]{Daeun Lee}
received the BS degree in electrical and computer engineering from Seoul National University (SNU), Seoul, Republic of Korea, in 2025. She is currently working toward the MS degree in computer science engineering at Seoul National University. Her research interests include 3D vision and computer graphics.
\end{IEEEbiography}
\vspace{-12mm}
\begin{IEEEbiography}[{\includegraphics[width=1in,height=1in,clip,keepaspectratio]{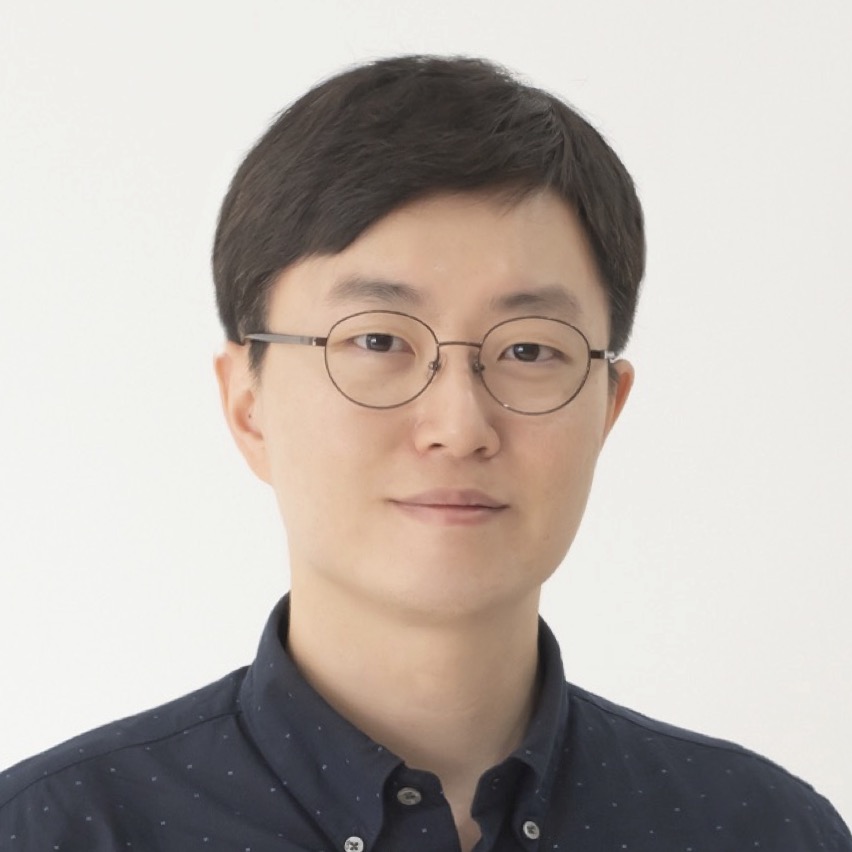}}]{Jaesik Park}
is an Associate Professor of Computer Science Engineering and an Interdisciplinary Program in AI at Seoul National University. He received his Bachelor’s degree from Hanyang University, and he received his Master’s and Ph.D. degrees from KAIST. He was a staff research scientist at Intel intelligent systems lab, where he co-created Open3D library. Before joining Seoul National University, he was a faculty member at POSTECH. His research interests include text-to-image synthesis, 3D perception, and computer vision topics. He serves as a program committee member at prestigious international conferences, such as CVPR, ECCV, ICCV, ICLR, ICML, ICRA, NeurIPS, and SIGGRAPH Asia.
\end{IEEEbiography}

\vfill

\clearpage
\appendices
\setcounter{page}{1}
\setcounter{section}{0}
\renewcommand{\thesection}{\Alph{section}}

\section{Overview}
\label{sec:supple_overview}
This supplementary material presents additional implementation details and results to support the main paper.
\begin{itemize}
    \item In Section \ref{sec:network}, we explain details of the \textit{Face-Fitting network} architecture and provide further implementation specifics.
    \item Section \ref{sec:visualization} shows the results of additional alignment results in controlled environments.
    \item Section \ref{sec:additional_results} shows the results of additional video results in uncontrolled environments.
\end{itemize}

\section{Implementation Details}
\label{sec:network}
\vspace{2mm}
Our unoptimized implementation runs offline, which can be boosted with parallel processing. When processing 100 frames of a single person in TBBT scenes in Sitcoms3D dataset~\cite{pavlakos2022one}, our pipeline takes about 30 minutes for stage reconstruction,  10 minutes for SMPL alignment, 1 hour for custom diffusion training, and 3 hours for actor reconstruction. We utilized a single NVIDIA A6000 GPU for training.

Spatio-temporal positioning module (Sec.~\ref{sec:positioning}) optimizes global translation $\textbf{t}$ and scale $s$ for the first 6k iterations using the total alignment loss (Eq.~\ref{eq:align_loss}). 
Since there are depth inconsistency between frames, we set $\lambda_{depth}$ in Eq.~\ref{eq:align_loss} as zero for the subsequent 2k iterations to ensure the smoothness of the 3D trajectory.

\begin{figure}[t]
    \centering
    \includegraphics[width=\linewidth]{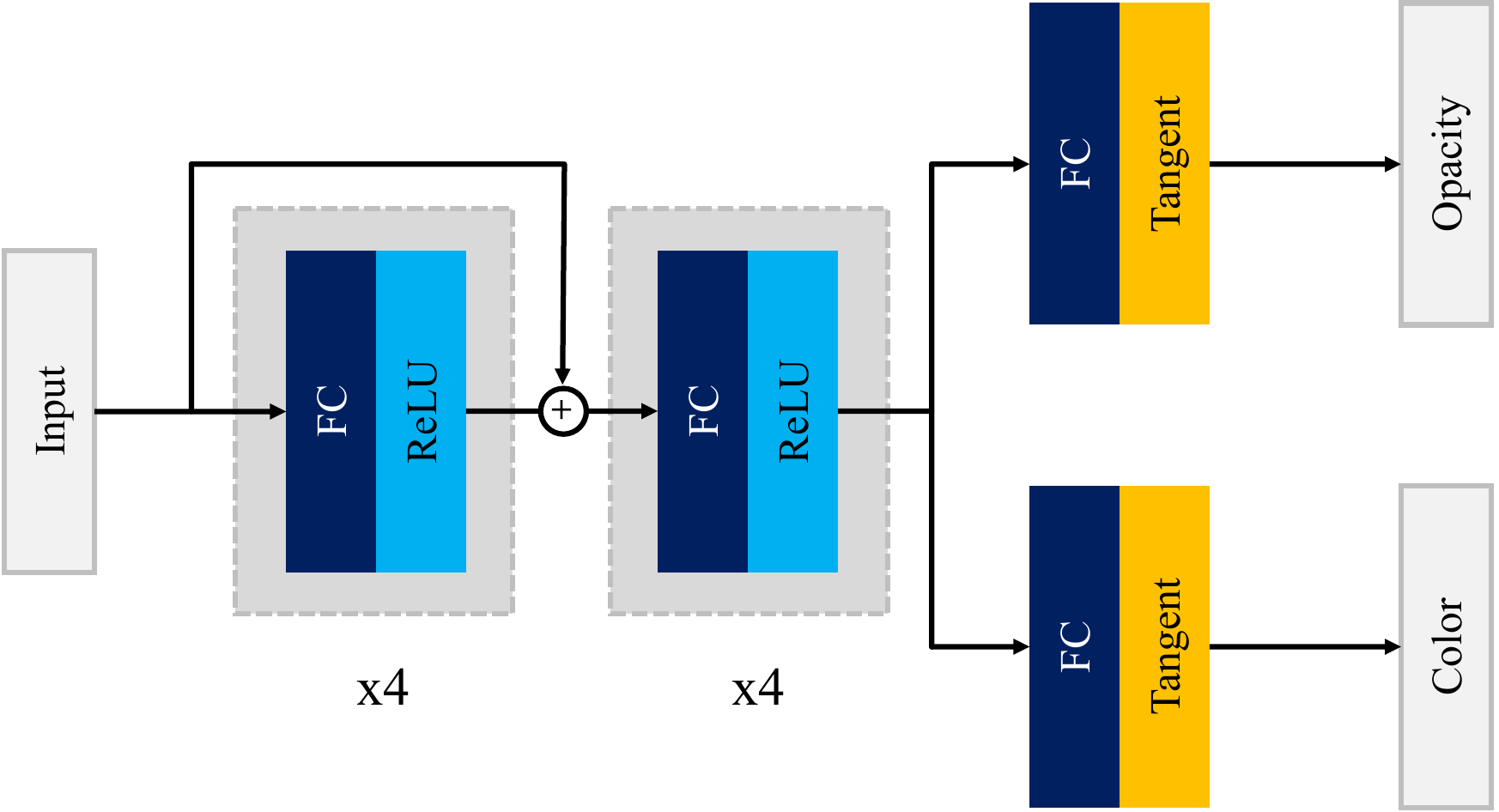}
    \caption{The architecture of our face fitting network.}
    \label{fig:network}
\end{figure}

\noindent\textbf{Face fitting network architecture.} Fig.~\ref{fig:network} shows the architecture of our face-fitting network (Sec.~\ref{sec:3D_actor}). We modify the deformation network from D-3DGS~\cite{jung2023deformable}. Instead of deforming the position, rotation, and scale of Gaussians, our network adjusts the color and opacity of Gaussians at each time step. In this way, our approach can capture the detailed expression change of the actors.

The face-fitting network takes Gaussian positions and time embeddings as input. To handle multiple actors, we concatenate each actor Gaussian set as $concat\{\mathcal{G}^{\text{actor}}_n\}^{N}_{n=1}$. Concatenated input is then processed through eight fully connected layers with ReLU activation functions. Additionally, the feature vector from the fourth layer is concatenated with the input. Output is a 256-dimensional feature vector, which is then passed to two separate fully connected layers. 
D-3DGS~\cite{jung2023deformable} does not utilize normalization at the end. However, since the opacity and color have a value between 0 and 1, we add a tangent hyperbolic activation function at the end to prevent overflow.

\section{Reconstruction Visualization}
\label{sec:visualization}

\begin{figure}[t]
    \centering
    \includegraphics[width=0.5\textwidth]{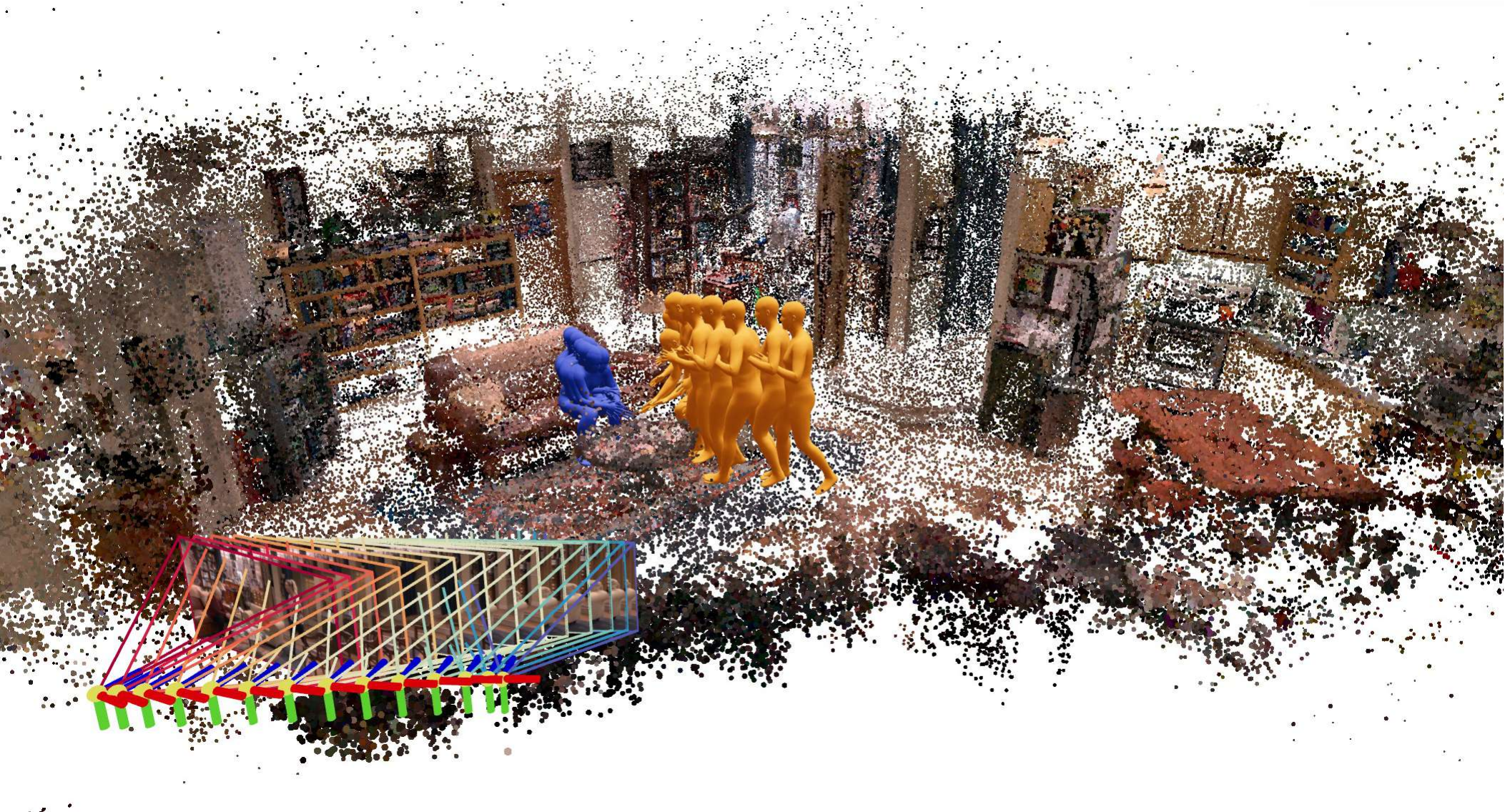}
    \vspace{-8mm}
    \caption{Visualization of \textbf{aligned actors}, \textbf{estimated cameras}, and \textbf{reconstructed 3D stage}.}
    \label{fig:actor_align}
    \vspace{-2mm}
\end{figure}

\begin{figure*}[b]
    \centering
    \includegraphics[width=0.91\textwidth]{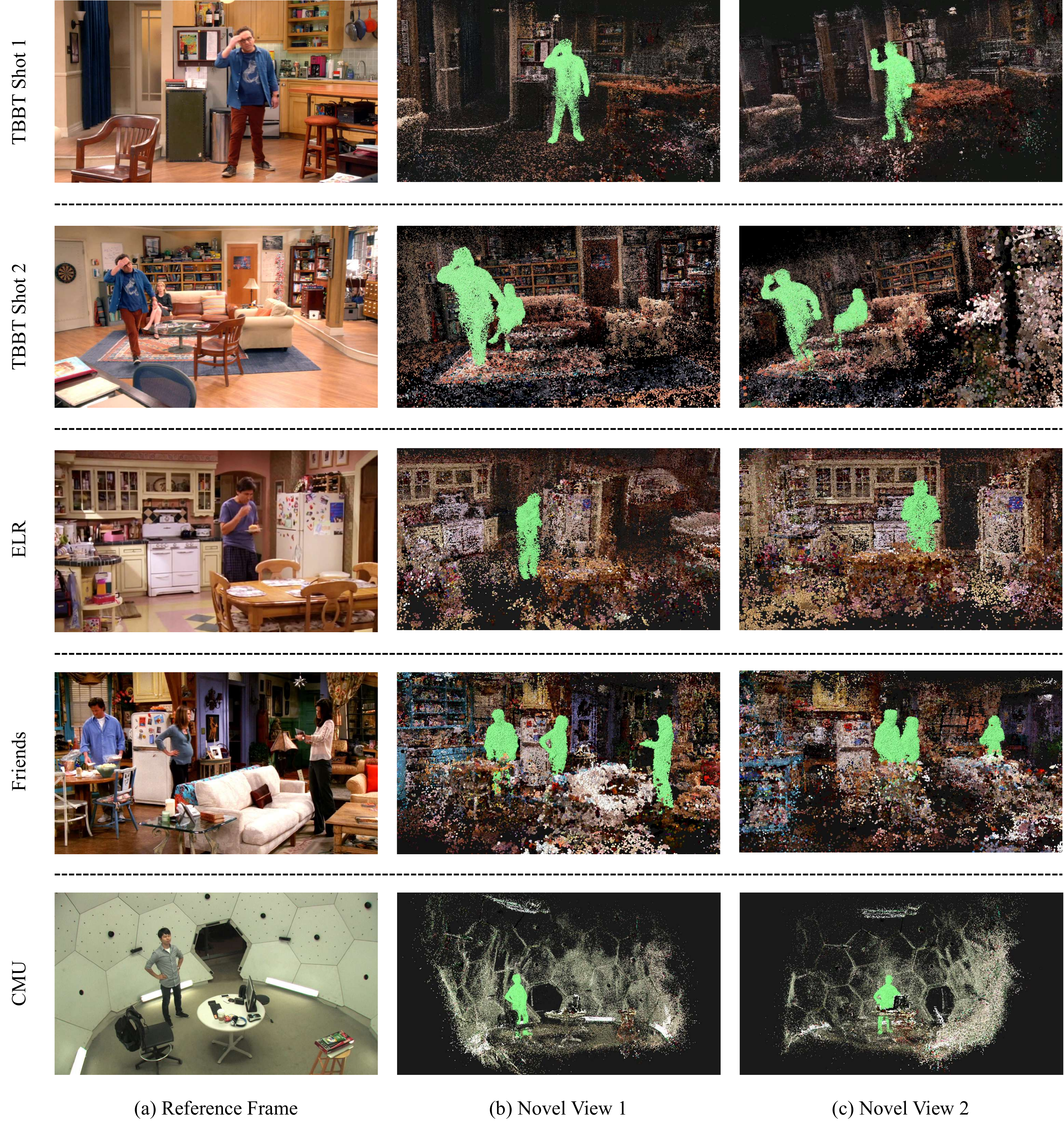}
    \vspace{-2mm}
    \caption{Additional results of the aligned actors in controlled environments. We visualize gaussian centers from two different novel viewpoints. Green points denote the Gaussian centers for the actors.}
    \label{fig:alignment}
    \vspace{10mm}
\end{figure*}

In this section, we present the visualization results of reconstruction process. As shown in Fig.~\ref{fig:actor_align}, by using \textit{spatio-temporal positioning} module, actors are correctly aligned to the stage. Additional results are given in Fig.~\ref{fig:alignment}. The green points indicate the centers of the actor Gaussians.

Unlike methods~\cite{guo2023vid2avatar, jiang2022neuman, kocabas2024hugs} that determine the scale of SMPL by identifying the intersection point between the ground plane and the feet, our approach optimizes scale using aligned depth information. This approach is robust to scenarios where actors are cropped or occluded by objects. 

\section{Uncontrolled Environments}
\label{sec:additional_results}
\begin{figure*}[t]
    \centering
    \includegraphics[width=0.91\textwidth]{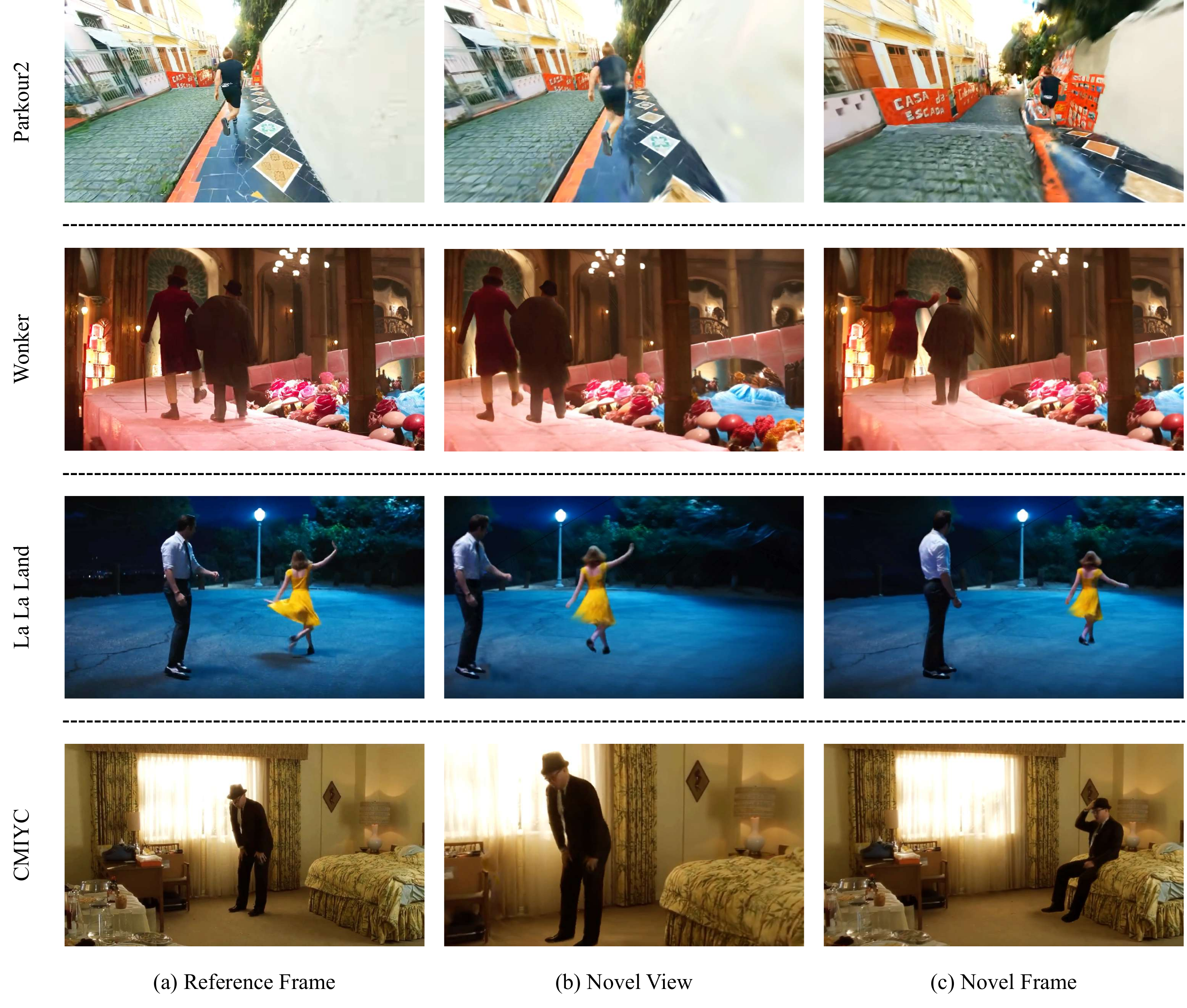}
    \vspace{-2mm}
    \caption{Additional results of the web video reconstruction. We select challenging web videos with dynamic human motions or fast camera movements. We render the scene from a novel viewpoint and a novel frame.}
    \label{fig:supp_web}
\end{figure*}

In this section, we present additional results from ShowMak3r++ on videos with uncontrolled environments. We select challenging web videos like 'Parkour', 'Wonker', 'La La Land', and 'Catch me if you can (CMIYC)', which have dynamic human motions or fast camera movements. After reconstruction, we render the scene from a novel viewpoint and a novel frame. As demonstrated in Fig.~\ref{fig:supp_web}, ShowMak3r++ shows robust performance even in uncontrolled environments. 

The following links correspond to the web videos used for this paper:  

\small{
\begin{itemize}
    \item \textbf{Joker:}\\ \texttt{\nolinkurl{www.youtube.com/watch?v=JeyVU4nMWCg}}
    \item \textbf{Ballet:}\\ \texttt{\nolinkurl{www.youtube.com/watch?v=zV1qLYukTH8}}
    \item \textbf{SNU dance:}\\ \texttt{\nolinkurl{www.instagram.com/reel/DNSpocSNg_M/?igsh=d3d0OGw1b2g1c3B0}}
    \item \textbf{Parkour1 and Parkour2:}\\ \texttt{\nolinkurl{www.youtube.com/watch?v=xbqVWZD-sfI}}
    \item \textbf{Wonker:}\\ \texttt{\nolinkurl{www.youtube.com/watch?v=zcctO0cYFLg}}
    \item \textbf{La La Land:}\\ \texttt{\nolinkurl{www.youtube.com/watch?v=_8w9rOpV3gc}}
    \item \textbf{CMIYC:}\\ \texttt{\nolinkurl{www.youtube.com/watch?v=eCWU3a4MhqI}}
\end{itemize}
}

\end{document}